\documentclass[twocolumn,journal]{IEEEtran}
\usepackage[square, comma, sort&compress, numbers]{natbib}
\usepackage{epsfig}
\usepackage{epstopdf}
\usepackage{citesort}
\usepackage{amssymb}
\usepackage{amsmath}
\usepackage{color}
\usepackage{url}
\usepackage{array}
\usepackage{multirow}
\usepackage{algorithm}
\usepackage{algorithmic}
\usepackage[table]{xcolor}
\usepackage{bm}
\usepackage{tablefootnote}
\usepackage{threeparttable}
\usepackage{color}
\usepackage{xcolor}
\usepackage{colortbl}
\usepackage{mdwmath,mdwtab,amsmath}
\usepackage{autobreak}
\usepackage{amsmath,amssymb}
\usepackage{array}
\usepackage{multirow}
\usepackage{tabularx}
\usepackage{makecell}
\usepackage{graphicx}
\usepackage{threeparttable}
\usepackage{tablefootnote}
\usepackage{hyperref}
\hypersetup{
	colorlinks=true,
	linkcolor=cyan,
	filecolor=lime,
	urlcolor=lime,
	citecolor=lime,
}
\newcolumntype{Y}{>{\centering\arraybackslash}X}

\newlength{\figurewidth}
\newlength{\smallfigurewidth}

\usepackage{tikz,xcolor,hyperref}
\definecolor{lime}{HTML}{A6CE39}
\DeclareRobustCommand{\orcidicon}{
	\begin{tikzpicture}
		\draw[lime, fill=lime] (0,0)
		circle[radius=0.16]
		node[white]{{\fontfamily{qag}\selectfont \tiny \.{I}D}}; 
	\end{tikzpicture}
	\hspace{-2mm}
}
\foreach \x in {A, ..., Z}{%
	\expandafter\xdef\csname orcid\x\endcsname{\noexpand\href{https://orcid.org/\csname orcidauthor\x\endcsname}{\noexpand\orcidicon}}
}

\begin{document}
	\title
	{Multi-Grained Angle Representation for \\Remote Sensing Object Detection}
	\author{%
		Hao Wang,
		Zhanchao Huang,
		Zhengchao Chen,
		\\
		Ying Song,
		and
		Wei Li,~\IEEEmembership{Senior Member,~IEEE}

		\thanks{%
			This work was supported by the National Key Research and Development Program of China 2021YFB3901103 and the Aeronautical Science Foundation of China under Grant 2020051072001. (Corresponding Author: Ying Song; e-mail: prisong@163.com. Wei Li; e-mail: liwei089@ieee.org)}
		\thanks{%
			Hao Wang, Zhanchao Huang are with the School of Information and Electronics, Beijing Institute of Technology, and Beijing Key Lab of Fractional Signals and Systems, 100081 Beijing, China. (e-mail: haohaolalahao@icloud.com, zhanchao.h@outlook.com).}
		\thanks{%
			Zhengchao Chen is with the Aerospace Information Research Institute, Chinese Academy of Sciences, 100094, Beijing, China. (e-mail: chenzc@radi.ac.cn).}
		\thanks{%
			Ying Song is with the School of information and communication engineering, Hubei University of Economics, Wuhan 430205, China. (e-mail: prisong@163.com).}
		\thanks{%
			Wei Li is with the School of Information and Electronics, Beijing Institute of Technology, Beijing 100811, China, and also with the Luoyang Institute of Electro-Optical Equipment, Aviation Industry Corporation of China, Ltd., Luoyang 471000, China. (e-mail: liwei089@ieee.org).}
	}
\maketitle
\thispagestyle{empty}
\pagestyle{empty}

\begin{abstract}
Arbitrary-oriented object detection (AOOD) plays a significant role for image understanding in remote sensing scenarios.
The existing AOOD methods face the challenges of ambiguity and high costs in angle representation.
To this end, a multi-grained angle representation (MGAR) method, consisting of coarse-grained angle classification (CAC) and fine-grained angle regression (FAR) , is proposed.
Specifically, the designed CAC avoids the ambiguity of angle prediction by discrete angular encoding (DAE) and reduces complexity by coarsening the granularity of DAE.
Based on CAC, FAR is developed to refine the angle prediction with much lower costs than narrowing the granularity of DAE.
Furthermore, an Intersection over Union (IoU) aware FAR-Loss (IFL) is designed to improve accuracy of angle prediction using an adaptive re-weighting mechanism guided by IoU.
Extensive experiments are performed on several public remote sensing datasets, which demonstrate the effectiveness of the proposed MGAR. Moreover, experiments on embedded devices demonstrate that the proposed MGAR is also friendly for lightweight deployments.

\end{abstract}


\section{Introduction}
\begin{figure}[htbp]
        \centering
        \includegraphics[width=1.0\linewidth]{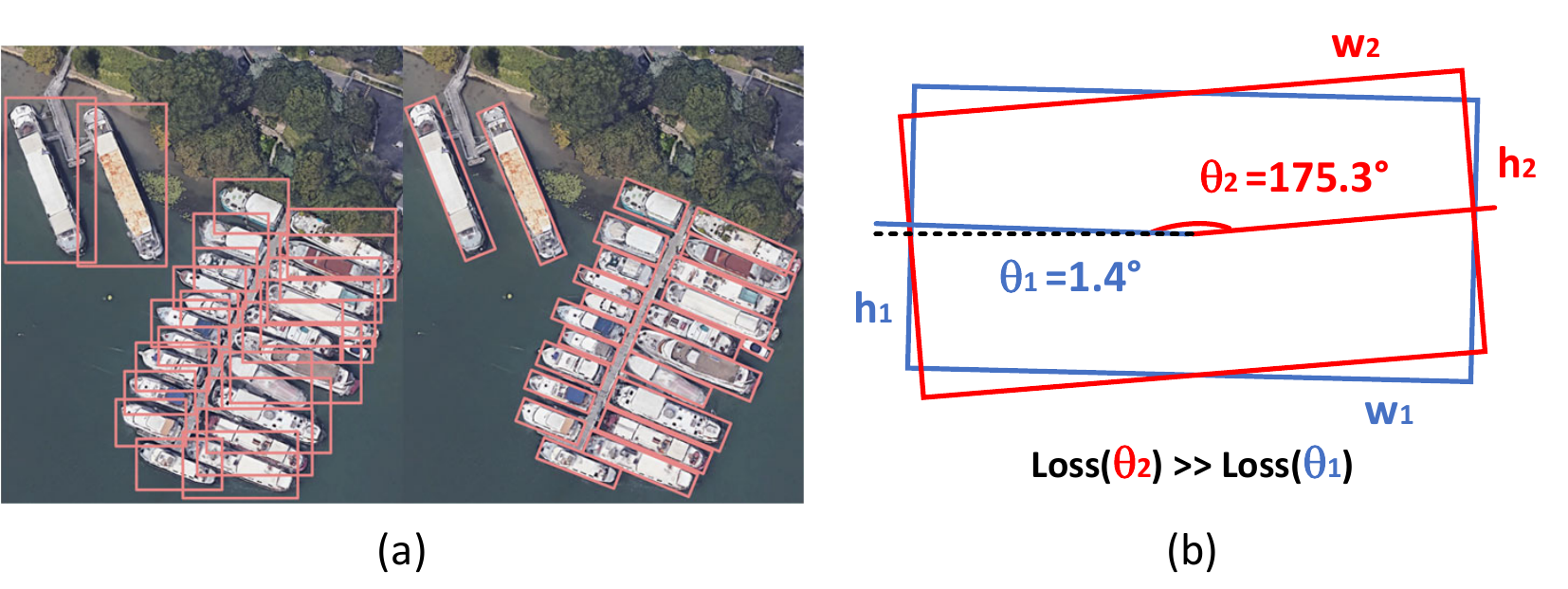}
        \caption{\textcolor{black}{(a) Different forms of annotation (Left: Horizontal bounding boxes (HBBs). Right: Oriented bounding boxes (OBBs)). (b) Mutation of regression loss.}}
        \label{fig0}
\end{figure}

\IEEEPARstart{I}{n} recent years, arbitrary-oriented object detection (AOOD)
has been widely used in complex remote sensing, aviation, and other scenes \cite{xia2018dota, li2020object, cheng2018learning}.
\textcolor{black}{The oriented bounding boxes (OBBs) provides richer angle information than the horizontal bounding boxes (HBBs)}, which helps to better localize objects at arbitrary angles in some particular scenes, such as ships arranged densely in the harbor, as shown in Fig. \ref{fig0} (a).

AOOD methods are developed from classical object detection algorithms, such as
Faster RCNN \cite{ren2015faster} and \textcolor{black}{RetinaNet} \cite{lin2017focal}.
The mainstream AOOD methods adopt five parameters to describe the position information of the OBB.
They are the coordinates $(x, y)$ of the OBB's center and OBB's width $w$, height $h$, and rotation angle $\theta$.
\textcolor{black}{Most of the existing AOOD methods directly predict the rotation angle by regression.
In the beginning, AOOD methods mainly improve the network structure to adapt to rotating objects \cite{han2021redet, qian2021learning, yang2021r3det}, and some Anchor-based methods \cite{ma2018arbitrary, 8438330} used rotated anchors rather than horizontal anchors to obtain more suitable regional proposals. Region of Interest (RoI) transformer \cite{ding2019learning} learns oriented proposals from horizontal RoI, and oriented Region-CNN (R-CNN) \cite{xie2021oriented} designs oriented region proposal network (RPN) to represent rotating proposals.
They all start with improving the method of proposals generation to obtain higher quality proposals.}
However, on the one hand, they may face ambiguous predictions because of the periodicity of angles.
\textcolor{black}{For instance, as shown in Fig. \ref{fig0} (b), the two angles of $1.4^{\circ}$ and $175.3^{\circ}$ are visually consistent,}
but the regression losses are different, which causes a sudden change in the loss function and further affects model learning.
On the other hand, the extensive regression range from $0^{\circ}$ to $180^{\circ}$ presents instability for the angle regression.
Although methods such as IoU-Smooth L1 loss \cite{yang2019scrdet} and Modulated loss \cite{qian2021learning} avoid loss mutation by adding constraints to the regression loss function to learn the angle representation better, they do not consider the problem from the fundamental perspective of angle regression.

\begin{figure}[htbp]
        \centering
        \includegraphics[width=1.0\linewidth]{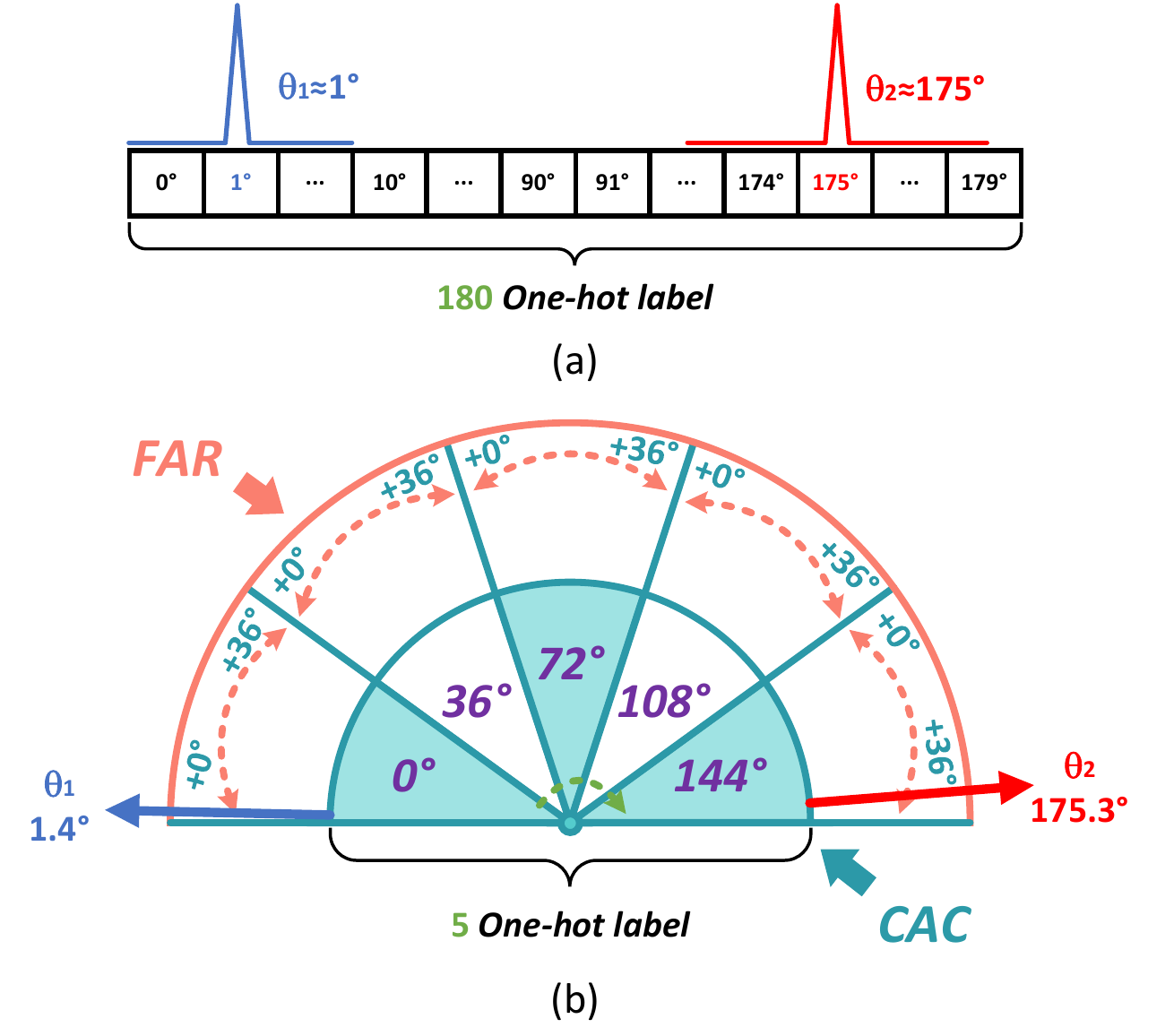}
        \caption{\textcolor{black}{Angle representation of the perspective based on the classification method. (a) Heavy prediction layers. (b) The proposed MGAR.}}
        \label{fig0_1}
\end{figure}
Circular Smooth Label (CSL) \cite{yang2020arbitrars} and Densely Coded Labels (DCL) \cite{yang2021dense} methods consider things from a new perspective, and they transform the angle regression to a discrete classification problem by converting continuous angle information to discrete angular encoding (DAE).
Although the CSL method resolves the ambiguity by fine-grained angle classification (FAC), which divides the angle into a category at 1-degree intervals in 0-180 degrees, \textcolor{black}{as shown in Fig. \ref{fig0_1} (a),} it still faces the following issues.
First, an additional hyperparameter, i.e., window size, is introduced by CSL to smooth one-hot encoding labels, but it is sensitive to different datasets.
In addition, FAC increases the number of prediction layers of the convolutional neural network (CNN) model, which affects the model's efficiency.
In this regard, DCL improves CSL by introducing Binary-Code and Gray-Code into FAC to obtain sparse encoding labels and reduce the CNN prediction layers. However, the encoding errors still exist, and the encoding length depends on a hyperparameter sensitive to different datasets.
In addition, the introduced label decoding also affects the inference speed of the CNN model.

To solve above issues, a multi-grained angle representation (MGAR) method is proposed.
The proposed MGAR divides the angle representation into two parts, \textcolor{black}{as shown in Fig. \ref{fig0_1} (b)}. The first part is coarse-grained angle classification (CAC), and the second part is fine-grained angle regression (FAR).
The CAC first divides the angle into multiple coarse categories and determines which category the angle belongs to. The FAR performs a fine continuous regression on a small range of angle classification categories.
Except for this, we design an Intersection over Union (IoU) aware FAR-Loss (IFL) for FAR to improve accuracy by using an adaptive re-weighting mechanism guided by IoU.
The main contributions of this paper are summarized as follows:

(1) The proposed MGAR includes CAC and FAR. The designed CAC refrains from the ambiguity of angle prediction, and increases model calculation efficiency by reducing angle categories.
In addition, the CAC is conducive to the learning of the classification task without introducing the hyperparameter sensitive to different datasets.

(2) Based on the CAC, the designed FAR makes the angle prediction more refined, and meanwhile, the encoding error is avoided, further improving the computing efficiency.
Furthermore, the IFL makes the angle regression smoother and more stable at convergence, achieving extremely high performance.


\textcolor{black}{(3) Experiments are implemented on five public large-scale remote sensing and aerial scene datasets. The results demonstrate the effectiveness of the proposed method in terms of accuracy and speed. Furthermore, the advantages of hyperparameter insensitivity and performance stability of the proposed MGAR are evaluated. The proposed MGAR can avoid hyperparameter selection and save training costs. Besides, the scalability of MGAR for lightweight deployment on embedding devices is analyzed. The proposed MGAR improves the performance of the lightweight AOOD model and maintains a fast detection speed under power-constrained conditions.
}

The rest of this paper is organized as follows. Section \uppercase\expandafter{\romannumeral2} introduces the related work and the existing angle representation methods for AOOD method. Section \uppercase\expandafter{\romannumeral3} introduces the proposed method MGAR. Section \uppercase\expandafter{\romannumeral4} analyzes the performance of the proposed MGAR through extensive experiments. Section \uppercase\expandafter{\romannumeral5} draws conclusions.

\section{Related Works}
\subsection{Arbitrary-Oriented Object Detection (AOOD)}
As a fundamental task of computer vision, object detection methods are widely
used in natural scene detection \cite{lin2014microsoft}, face recognition \cite{deng2020retinaface}, and other fields \cite{wu2019orsim, wu2019fourier}. Current
CNN-based object detection methods are divided into two-stage
methods and single-stage methods from the network structure. The RCNN series \cite{girshick2014rich, ren2015faster} accelerate the development of two-stage methods but are limited by the detection speed and model size. Therefore, single-stage methods that focus more on speed are proposed, such as RetinaNet \cite{lin2017focal} and YOLO series \cite{redmon2016you, redmon2017yolo9000, redmon2018yolov3}. In recent years, object
detection methods are different from the past Anchor-based methods, and a
series of Anchor-free methods have emerged successively, such as Fully Convolutional One-Stage (FCOS) object detection \cite{tian2019fcos} and CornerNet \cite {law2018cornernet}, etc. AOOD methods are
mainly derived from object detection methods. Some AOOD methods have been developed to deal with the challenges of various angle information in remote sensing, aviation and other special scenes. The two-stage method SCRDET \cite {yang2019scrdet}
introduces the attention module to eliminate the noise problem caused by
complex remote sensing scenes. The single-stage method R3Det \cite{yang2021r3det}
designs a feature refinement module based on RetinaNet to further refine the
OBB. Anchor-free method General Gaussian Heatmap Label (GGHL) \cite{9709203} further refines the features of
the rotating object by introducing the two-dimensional Gaussian label
distribution based on FCOS. In addition, there is also a method named
LO-Det \cite{9390310} for lightweight research on AOOD.

\begin{figure}[htbp]
        \centering
        \includegraphics[width=0.7\linewidth]{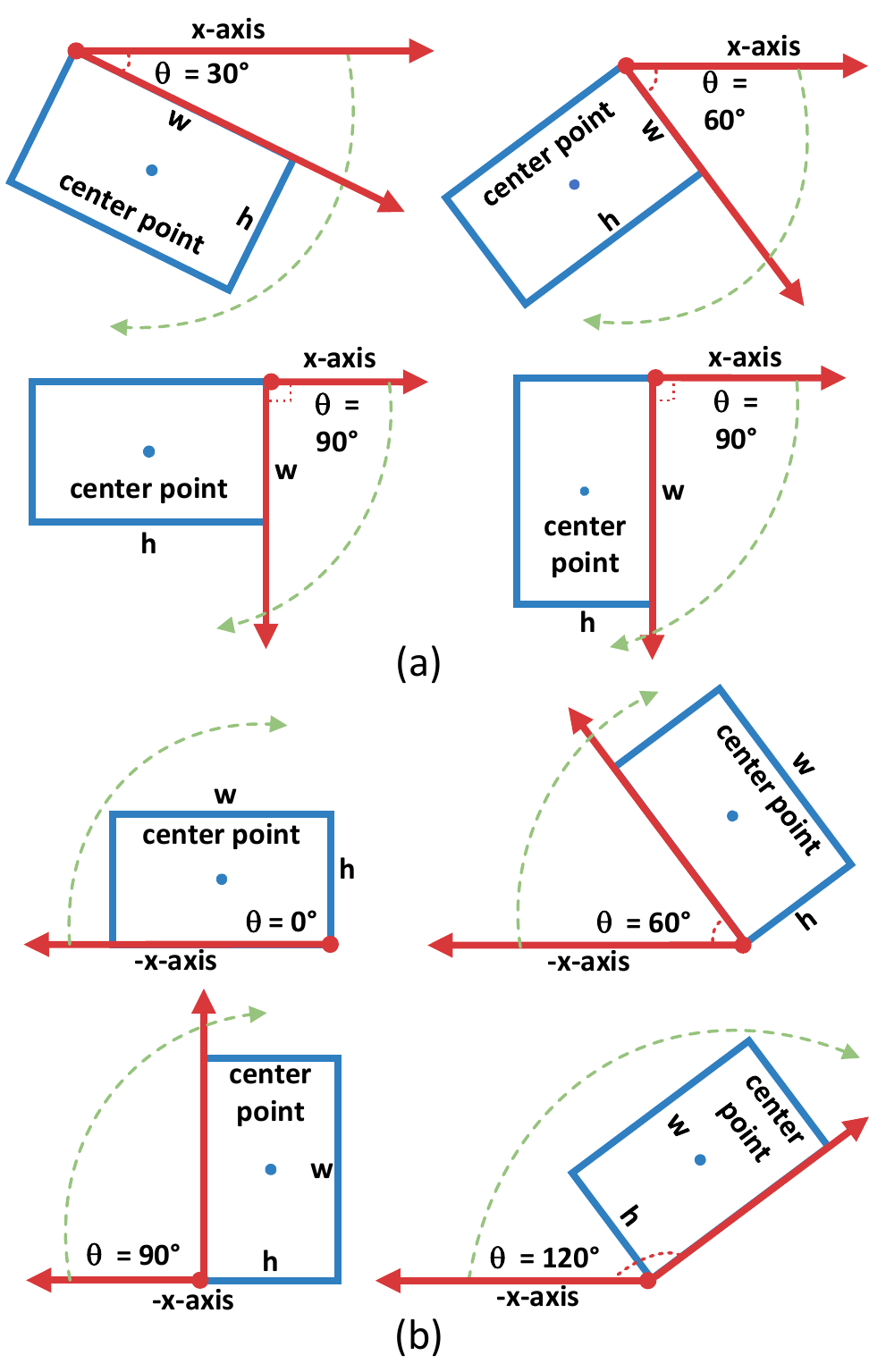}
        \caption{Two five-parameter methods. (a) OpenCV-based method with 90° angle range. (b) Long side based method with 180°.}
        \label{fig1}
\end{figure}

\subsection{Angle Representation of OBB}
The definition of the OBB in existing AOOD methods is divided into two categories: the five-parameter method and the eight-parameter
method, where the five-parameter method is divided into the 90° definition
method based on the OpenCV and the $180^{\circ}$ definition method extended on this
definition. There are many problems in using the regression-based method to
predict angles directly. Some methods solve the problems in OBB regression
from the loss function, such as SkewIOU \cite{ma2018arbitrary}, which is optimized for the large
aspect ratio problem. Some methods, such as Gaussian  Wasserstein Distance (GWD) \cite{yang2021rethinking},
convert the five-parameter method to a two-dimensional Gaussian distribution
representation and design a novel loss function to regress the OBB indirectly.
Other methods such as CSL and DCL utilize the idea of classification to
predict angles.
The eight-parameter method uses the four coordinate position representation of the OBB.
However, there is a vertex
sorting problem. Gliding Vertex \cite{xu2020gliding} avoids sorting by
changing the representation of the bounding box, while RSDet \cite{qian2021learning} designs corner points sorting algorithm to achieve OBB prediction.

The current AOOD methods have different definitions for the five-parameter
method. For the OpenCV, as shown in Fig. \ref{fig1}(a), \textcolor{black}{the $\theta$ is defined as the angle between the positive x-axis and the first side it encounters when rotating clockwise, and the angle ranges from $(0^{\circ}, 90^{\circ}$]}. In this definition, there is a possibility of exchange between the two sides of the
OBB, and the angle range varies periodically. These two problems lead to abrupt
changes and discontinuities in the model's loss function during training. To
avoid the problems caused by the OpenCV-based five-parameter method, we adopt
the $180^{\circ}$ definition method based on the long side. It is displayed in
Fig. \ref{fig1} (b), where the longest side is $w$ and the shortest side is $h$.
Furthermore, the angle is defined as the angle between the long side of the OBB and the negative x-axis, and the angle range is \textcolor{black}{$[0^\circ, 180^\circ)$}.

\section{Proposed Detection Methodology}
\begin{figure}[htbp]
        \centering
        \includegraphics[width=1.0\linewidth]{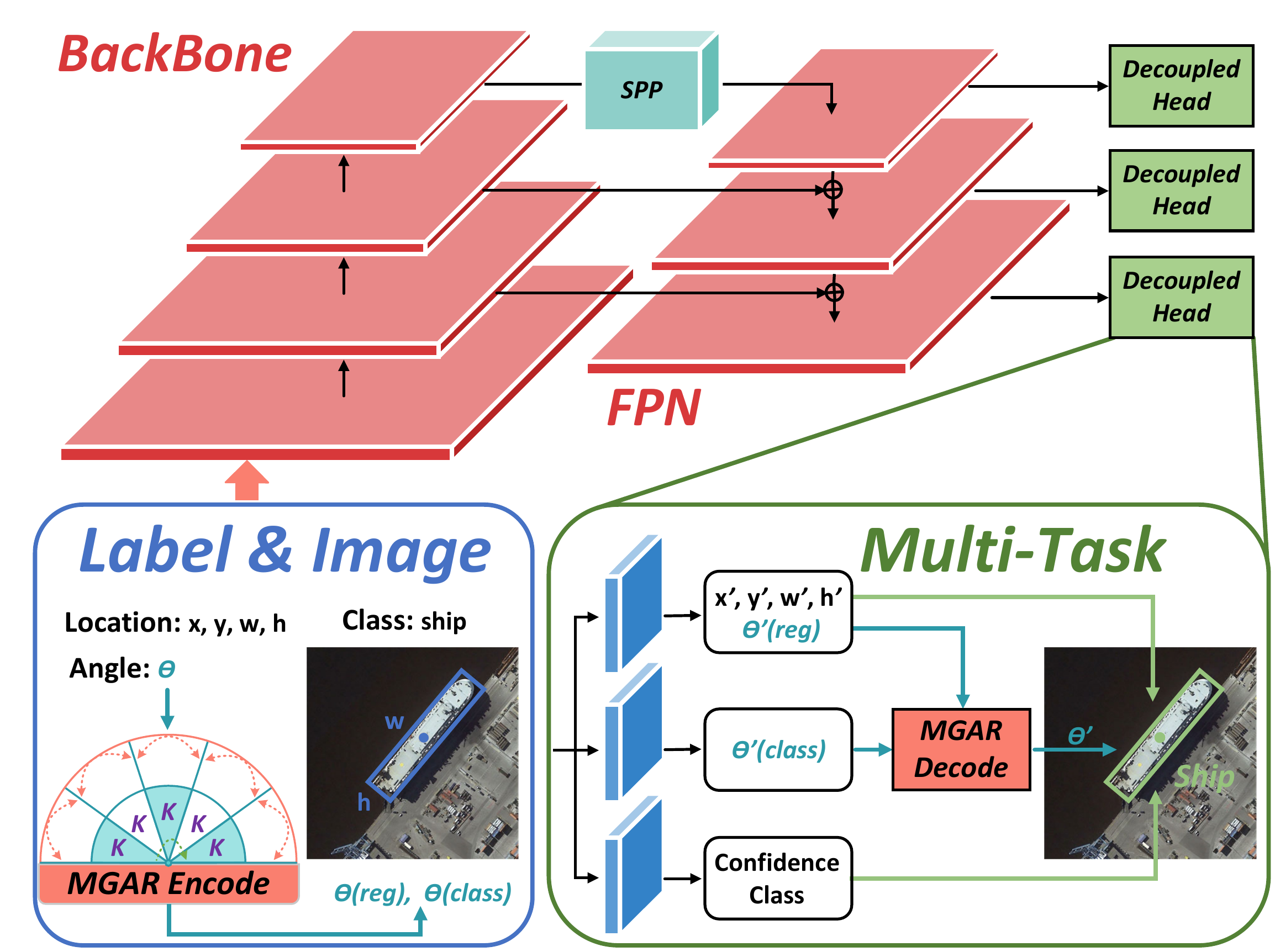}
        \caption{The baseline framework architecture for MGAR method. Including three parts: BackBone, FPN and Decoupled Head. The decoupled head contains three multi-task branches which are the location regression branch, angle classification branch, and object classification branch.}
        \label{fig2}
\end{figure}
Fig. \ref{fig2} illustrates the baseline framework for the proposed MGAR, including BackBone, Feature Pyramid Networks (FPN) \cite{lin2017feature}, and Decoupled Head.
Then the proposed MGAR method is compared with the regression-based, CSL,
and DCL methods in four aspects: representation, error analysis, prediction
layer thickness, and hyperparameters. The loss function is introduced in the
end.
\subsection{Baseline Framework}
Compared to the two-stage detection methods, the single-stage detection methods achieve a better balance between speed and accuracy. They are more convenient when designing a lighter model for practical deployment. Therefore we adopt the widely used single-stage method YOLO as the baseline and improved it for the AOOD task. To avoid the impact of complex networks on the performance of the proposed MGAR, we choose the YOLOv3 \cite{redmon2018yolov3} model with a more concise network structure to adapt to the AOOD better and use this model as the baseline. The mainstream single-stage models usually consist of three parts: backbone, neck, and head. Our baseline model improved from YOLO utilizes Darknet53 \cite{redmon2018yolov3} as the backbone and FPN as the neck and adopts a highly coupled detection head. We add the Spatial Pyramid Pooling (SPP) \cite{he2015spatial} structure between the backbone and FPN to achieve better feature fusion. The coupled head is transformed into a decoupled head to avoid the impact of an overly coupled prediction layer on different prediction parameters. In addition, the output of each FPN layer is decomposed into three multi-task prediction branches: regression branch, class classification branch, and angle classification branch. The regression branch is responsible for regressing the center coordinates of the OBB, the long side $w$, the short side $h$, as well as the regression part of the angle information. The class classification branch predicts the category to which the object belongs and distinguishes the confidence score between the foreground and the background. The angle classification branch is employed to predict the class of the angle information.
\begin{figure*}[htbp]
        \centering
        \includegraphics[width=0.8\textwidth]{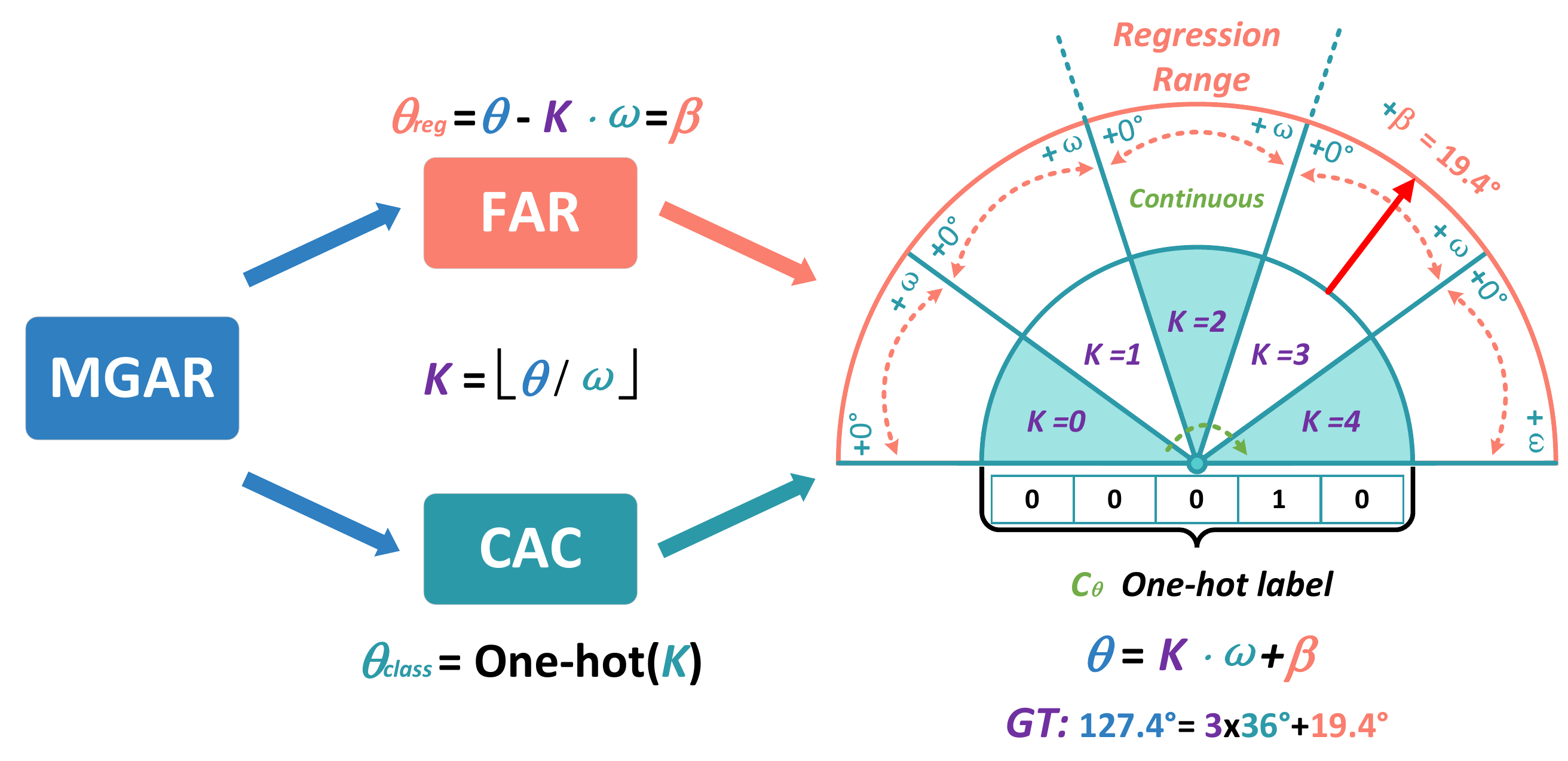}
        \caption{The proposed MGAR is divided into two parts: the CAC and the FAR. The CAC encodes angle classification label and the FAR encodes angle regression label. A specific example is illustrated.}
        \label{fig3}
\end{figure*}
\subsection{Multi-Grained Angle Representation (MGAR)}
The proposed MGAR divides the angle information into two parts for
representation. One is the coarse-grained classification part denoted as
$\theta_{class}$; the other is the fine-grained regression part expressed as
$\theta_{regression}$. For the prediction of the network, the proposed MGAR has
an encoding and decoding processes. We use $\theta_{gt}$ to denote the true angle
label in the range of \textcolor{black}{$[0^{\circ}, 180^{\circ})$}, $\theta_{encodeClass}$ to denote the angle
encoding classification label, $\theta_{encodeRegression}$ to indicate the angle
encoding regression label, and $t_{\theta_{class}}^{'}$ to denote the prediction
result of the angle classification part of the network, and
$t_{\theta_{reg.}}^{'}$ to denote the prediction of the angle regression
part of the network. Correspondingly, the decoding information is represented
by $\theta_{decodeClass}$ and $\theta_{decodeRegression}$. The final angle
prediction result is expressed by $\theta^{'}$. The specific encoding and decoding processes
are represented as follows:
\begin{equation}
\textcolor{black}
{
        k=\lfloor\frac{\theta_{gt}}{\omega}\rfloor
}
\end{equation}
\begin{equation}
\textcolor{black}
{
        \theta_{encodeClass}= Onehot(k)
}
\end{equation}
\begin{equation}
\textcolor{black}
{
        \theta_{encodeRegression}= \theta_{gt}-k\times\omega
}
\end{equation}
\begin{equation}
        \theta_{decodeClass}=\omega \times Argmax(Sigmoid(t_{\theta_{class}}^{'}))
\end{equation}
\begin{equation}
\textcolor{black}
{
        \theta_{decodeRegression} = {(t_{\theta_{reg.}}^{'})}^{2}
}
\end{equation}
\begin{equation}
        \theta^{'} =\theta_{decodeClass}+\theta_{decodeRegression}
\end{equation}
where $\omega=AR/C_{\theta}$ denotes the discretization granularity of the angle classification part. $AR$ represents the angle range, which is $180^{\circ}$.
$C_{\theta}$ indicates the number of coarse-grained categories that the angle needs to be divided into.
\textcolor{black}{$k$ represents the specific category of the angle.}
We use $Square$ function to fitting $t_{\theta_{reg.}}^{'}$ which is smoother than $Linear$ function.
The fine-grained regression range of the angle is \textcolor{black}{$[0^{\circ}, \omega)$}. The detailed MGAR is displayed in Fig. \ref{fig3}.

\subsection{Coarse-Grained Angle Classification (CAC)}
In the CAC, when $C_{\theta}=1$, the number
of angle classification is only one category, and the angle regression range
is \textcolor{black}{$[0^{\circ}, 180^{\circ})$}. At this time, MGAR is converted to a regression-based
method. When $C_{\theta}=180$, the angle regression range is \textcolor{black}{$[0^{\circ}, 1^{\circ})$}, and the
angle classification part is equivalent to the classification-based angle
encoding method CSL. The value range of $C_{\theta}$ is $[1, 180]$ theoretically,
but two problems need to be considered in practical applications. The first
problem is that when $C_{\theta}$ is not divisible by 180, $\omega$ is
not an integer, which makes the angle classification branch produce
an unavoidable floating-point calculation error when decoding. In order to avoid
this error, $C_{ \theta}$ should be the number of categories that can be
divided by 180. The second problem is that when the classification granularity
is too fine, such as in the case of $C_{\theta}=180$, there are too many
classification categories, increasing the difficulty of model training. CSL
introduces the Gaussian window function based on the one-hot label, converting
hard label to soft label. Although CSL overcomes the hindrance of fine-grained
classification in learning, it also introduces a dataset-sensitive
hyperparameter window size, which needs to be adjusted for different datasets.
Our method utilizes CAC.
Specifically, the introduced hyperparameter $C_{\theta}$ is insensitive to the datasets and $C_{\theta} \in [3, 4, 5]$.
When the number of categories is small, the one-hot label can be used to obtain a
good enough classification result without adding additional window functions.

The DCL method reduces the length of angle classification labels through
binary encoding and gray encoding. However, the introduced hyperparameter $C_ {\theta}$ is
sensitive to the dataset, and the encoding error can not be ignored. The
maximum encoding error of the two encoding methods is represented as follows:
\begin{equation}
        MAX(error)=\frac{\omega}{2}=\frac{90}{C_{\theta}}
\end{equation}
The average encoding error is represented as follows:
\begin{equation}
        E(error)=\int_{a}^{b}\frac{x}{b-a}dx=\int_{0}^{\frac{\omega}{2}}\frac{x}{\frac{\omega}{2}-0}dx=\frac{\omega}{4}=\frac{45}{C_{\theta}}
\end{equation}

\begin{table}[htbp]
        \centering
        \caption{Comparison of four angle representation methods, including maximum encoding error, average encoding error, introduced hyperparameter and its sensitivity}
        \label{table1}
        \resizebox*{\linewidth}{!}{
                \setlength{\tabcolsep}{1.0mm}{
                        \renewcommand\arraystretch{1.3}{
\textcolor{black}{
\begin{tabular}{c|c|cc|cc}
\hline\hline
Methods & $C_\theta$ & \begin{tabular}[c]{@{}c@{}}Maximum \\Encoding Error\end{tabular} & \begin{tabular}[c]{@{}c@{}}Average \\Encoding Error\end{tabular} & Hyperparameter & Sensibility  \\
\hline
Reg.    & 1          & 0                                                                & 0                                                                & -              & -            \\
CSL     & 180        & 0.5                                                              & 0.25                                                             & Window Size    & yes          \\
DCL     & 256        & 0.3515625                                                       & 0.17578125                                                      & $C_\theta$     & yes          \\
MGAR    & 3          & 0                                                                & 0                                                                & $C_\theta$     & No           \\
\hline\hline
                                \end{tabular}
}
                        }}}
\end{table}
The specific error profiles of the methods are listed in Table \ref{table1}.
The proposed MGAR does not produce errors when encoding angles. Specifically,
MGAR encodes continuous angle information into the regression part, avoiding
the different errors caused by the angle encoding method based only on the
classification method when encoding $\theta_{gt}$.

\subsection{Fine-Grained Angle Regression (FAR)}
The FAR of MGAR relies on the CAC to reduce the range from the original
\textcolor{black}{$[0^{\circ}, 180^{\circ})$} to \textcolor{black}{$[0^{\circ}, \frac{180^{\circ}}{C_{\theta}})$}. Narrowing of the angle regression
range significantly improves detection accuracy and reduces the volatility of the regression. Meanwhile, the regression value range is continuous, which
regresses the angle information more finely while avoiding the problem of
loss mutation. Furthermore, it fits the OBB more accurately than the
classification-based method only.

The thickness of the network's final prediction layer is inconsistent under
different angle representation methods. We use $Th$ to denote the prediction layer thickness and $A$ to represent the number of anchors. For CSL,
the prediction layer thickness is expressed as follows:
\begin{equation}
        Th_{CSL} = A \times AR / \omega = A \times C_{\theta}
\end{equation}
where $C_{\theta} = 180$.

For DCL, the prediction layer thickness is expressed as follows:
\begin{equation}
        Th_{DCL} = A \times [log_{2}(AR/\omega)] = A \times [log_{2}(C_{\theta})]
\end{equation}
where $C_{\theta} \in [32, 64, 128, 256]$.

For the proposed MGAR, the prediction layer thickness is expressed as
follows:
\begin{equation}
        Th_{MGAR} = A \times (AR/\omega + 1) = A \times (C_{\theta} + 1)
\end{equation}
where $C_{\theta} \in [3, 4, 5]$.

For the baseline framework, when $C_{\theta}$ taking the minimum value allowed by each
method, the prediction layer thickness, flops, and parameters of
several encoding methods are listed in Table \ref{table2}.

\begin{table}[htbp]
        \centering
        \caption{Comparison of three methods for prediction layer thickness, FLOPs(G) and Parameters(M), under the same baseline framework.}
        \label{table2}
        \begin{threeparttable}
                \setlength{\tabcolsep}{1.0mm}{
                        \renewcommand\arraystretch{1.3}{
\textcolor{black}{
\begin{tabular}{c|c|c|c|c|c}
\hline\hline
Methods        & $C_\theta$ & Anchor & Thickness   & FLOPs(G)          & Parameters(M)     \\
\hline
baseline+CSL  & 180        & 9      & 1620        & 141.8064          & 75.9356           \\
baseline+DCL  & 32         & 9      & 45          & 139.4475          & 75.9356           \\
baseline+MGAR & 3          & 9      & \textbf{36} & \textbf{139.4340} & \textbf{74.9878}  \\
\hline\hline
\end{tabular}
}
                        }}
                        \begin{tablenotes}
                        \footnotesize
                        \item {Note: The unit G is Giga, which represents $1\times10^{9}$. The unit M represents $1\times10^{6}$.}
                        \end{tablenotes}
                \end{threeparttable}
\end{table}
The proposed MGAR requires about 97\% less prediction layer thickness
and about 2.37 fewer \textcolor{black}{floating-point operations (FLOPs)} for angle representation than CSL. The required
prediction layer thickness for MGAR is comparable to DCL. Nevertheless the DCL method
introduces binary and gray encoding, resulting in additional time overhead in
decoding the angle information. Overall, the proposed MGAR method helps to reduce the
computational effort and improve the computational efficiency of the model,
especially in lightweight deployment.

\subsection{Loss Function}
For the proposed method, we use six parameters $(x,y,w,h,\theta_{class},
        \theta_{reg.})$ to represent the OBB, where $(\theta_{class},
        \theta_{reg.})$ represents the two parts of $\theta$, $(x, y)$ denotes
the relative coordinates of the center point of the OBB, and $(w,h)$
corresponds to the long and short sides of the OBB, respectively. For the other
five parameters to be regressed, the regression equations are indicated as
follows:
\begin{equation}
        t_{x} = (x-x_{a})/w_{a}, \quad t_{y} = (y-y_{a})/h_{a}
\end{equation}
\begin{equation}
        t_{w} = log(w/w_{a}), \quad t_{h} = log(h/h_{a})
\end{equation}
\begin{equation}
\textcolor{black}{
        t_{\theta_{reg.}} = \sqrt{\theta_{reg.}}
}
\end{equation}
\begin{equation}
        t_{x}^{'} = (x^{'}-x_{a})/w_{a}, \quad t_{y}^{'} = (y^{'}-y_{a})/h_{a}
\end{equation}
\begin{equation}
        t_{w}^{'} = log(w^{'}/w_{a}), \quad t_{h}^{'} = log(h^{'}/h_{a})
\end{equation}
\begin{equation}
\textcolor{black}{
        t_{\theta_{reg.}}^{'} = \sqrt{\theta_{reg.}^{'}}
}
\end{equation}
where $(x_{a}, y_{a}, w_{a}, h_{a})$ denotes the center coordinates, long side, and short side of the anchor box, respectively. \textcolor{black}{$(x, y, w, h, \theta_{reg.})$} denotes the ground truth OBB, and \textcolor{black}{$(x^{'}, y^{'}, w^{'}, h^{'}, \theta_{reg.}^{'})$} represents the final prediction value. \textcolor{black}{$(t_{x}, t_{y}, t_{w}, t_{h}, t_{\theta_{reg.}})$} is the final output value of the network.

The multi-task loss function consists of five components: location regression,
confidence classification, class classification, angle category
classification, and angle regression. The specific loss function used in the proposed MGAR is expressed as follows:
\begin{equation}
        \begin{gathered}
                L=\frac{\lambda_{1}}{N} \sum_{n=1}^{N}obj_{n} \times L_{IoU}\left(\left(x^{\prime}, y^{\prime}, w^{\prime}, h^{\prime}\right),(x, y, w, h)\right) \\
                +\frac{\lambda_{2}}{N} \sum^{N} L_{conf}\left({conf}^{\prime}, {conf}\right)
                +\frac{\lambda_{3}}{N} \sum_{n=1}^{N} obj_{n} \times L_{cls}(p, t) \\
                +\frac{\lambda_{4}}{N} \sum_{n=1}^{N} obj_{n} \times L_{cls}\left(\theta_{{class}}^{\prime},  \theta_{{class}}\right) \\
                +\frac{\lambda_{5}}{N} \sum_{n=1}^{N} obj_{n} \times L_{{reg}}\left(\theta_{{reg. }}^{\prime}, \theta_{{reg. }}\right) \times (|-log(IoU)|+1)
        \end{gathered}
\end{equation}
where hyperparameter $\lambda_{i}(i=1,2,3,4,5)$ controls the weight distribution of different loss components, \textcolor{black}{for all datasets, we set them to $(2,2,5,2,0.5)$ respectively.} The $obj_n$ serves to distinguish whether the assigned label is foreground or background. It represents foreground when $n=0$ and represents background when $n=0$ . For the four parameters $(x, y, w, h)$, we use GIoU loss \cite{rezatofighi2019generalized} for regression. The $conf$ denotes the object confidence, which predicts the probability that the objects belong to the foreground, and we use Focal loss \cite{lin2017focal} to calculate this loss. For the class of the objects, we regard $p$ as the predicted class probability and $t$ as the true class, and we adopt the cross-entropy function to calculate the class loss.
The same cross-entropy is used for the coarse-grained classification of the angles to learn the encoding angle category information.
For the angle regression, we design an IoU-aware FAR-Loss (IFL) , which introduces the object's IoU score based on smooth L1 loss \cite{ren2015faster} and uses $|-log(IoU)|+ 1$ to adaptively re-weighting the loss. IFL can guide angle regression more smoothly.

\section{Experiments and Analysis}
In this section, experiments on public remote sensing datasets are conducted to
verify the effectiveness of the proposed MGAR.
The hardware and software platforms, implementation details, and experimental datasets are presented first.
Then ablation experiments are performed, comparative experiments and discussions are carried out on several public datasets.

All the experiments are implemented on a computer with an AMD EPYC 7542@2.9GHz
CPU, 128GB of memory, and two GPUs of NVIDIA GeForce RTX 3090 24GB. In addition,
to verify that the proposed method is conducive to lightweight deployment, we
also test it on the embedded device NVIDIA Jetson AGX Xavier. We use Pytorch \cite{https://doi.org/10.48550/arxiv.1912.01703} to
implement the framework.

\begin{table}[htbp]
\centering
\caption{Ablation studies evaluated under mAP(\%)(VOC12) on the HRSC2016 dataset}
\label{table3_1}
\begin{threeparttable}
\resizebox*{\linewidth}{!}{
\setlength{\tabcolsep}{1.0mm}{
\renewcommand\arraystretch{1.3}{
\textcolor{black}{
\begin{tabular}{c|c|c|ccc|c}
\hline\hline
Methods                                                         & \multicolumn{1}{c|}{SPP}  & $C_\theta$ & mAP$_{50}$                                                               & mAP$_{85}$                                                                & mAP$_{50:95}$                                                             & \begin{tabular}[c]{@{}c@{}}Speed\\(FPS)\end{tabular}                      \\
\hline
Regression                                                             & \multicolumn{1}{c|}{}     & 1          & 91.42                                                                        & 11.57                                                                         & 49.99                                                                         & 51.73                                                                        \\
\hline
Reg. (baseline)                                                  & $\checkmark$ & 1          & 92.02                                                                    & 14.97                                                                     & 52.86                                                                     & 51.44                                                                     \\
\hline
\begin{tabular}[c]{@{}c@{}}baseline\\+ CSL\end{tabular}         & $\checkmark$ & 180        & \begin{tabular}[c]{@{}c@{}}97.41\\(+5.39)\end{tabular}                   & \begin{tabular}[c]{@{}c@{}}43.72\\(+28.75)\end{tabular}                   & \begin{tabular}[c]{@{}c@{}}68.35\\(+15.49)\end{tabular}                   & \begin{tabular}[c]{@{}c@{}}50.53\\(-0.91)\end{tabular}                    \\
\hline
\begin{tabular}[c]{@{}c@{}}baseline\\+ DCL(binary)\end{tabular} & $\checkmark$ & 128        & \begin{tabular}[c]{@{}c@{}}93.86\\(+1.84)\end{tabular}                   & \begin{tabular}[c]{@{}c@{}}18.06\\(+3.09)\end{tabular}                    & \begin{tabular}[c]{@{}c@{}}56.23\\(+3.37)\end{tabular}                    & \begin{tabular}[c]{@{}c@{}}52.63\\(+1.19)\end{tabular}                    \\
\hline
\begin{tabular}[c]{@{}c@{}}baseline\\+ DCL(gray)\end{tabular}   & $\checkmark$ & 64         & \begin{tabular}[c]{@{}c@{}}97.44\\(+5.42)\end{tabular}                   & \begin{tabular}[c]{@{}c@{}}42.09\\(+27.12)\end{tabular}                   & \begin{tabular}[c]{@{}c@{}}68.18\\(+15.32)\end{tabular}                   & \begin{tabular}[c]{@{}c@{}}53.24\\(+1.80)\end{tabular}                    \\
\hline
\begin{tabular}[c]{@{}c@{}}baseline\\+ MGAR\end{tabular}        & $\checkmark$ & 5          & \begin{tabular}[c]{@{}c@{}}\textbf{97.62}\\\textbf{(+5.60)}\end{tabular} & \begin{tabular}[c]{@{}c@{}}\textbf{49.58}\\\textbf{(+34.61)}\end{tabular} & \begin{tabular}[c]{@{}c@{}}\textbf{68.83}\\\textbf{(+15.97)}\end{tabular} & \begin{tabular}[c]{@{}c@{}}\textbf{56.21}\\\textbf{(+4.77)}\end{tabular}  \\
\hline\hline
\end{tabular}
}
}}}
\begin{tablenotes}
\footnotesize
\item {Note: Speed is the test result on NVIDIA GeForce RTX 3090. The speed (average of 10 tests) includes the network inference speed with post-processing. The input size of image for network is 800$\times$800 pixels.}
\end{tablenotes}
\end{threeparttable}
\end{table}

\begin{table}
\centering
\caption{\textcolor{black}{Comparison of Hyperparameter's sensitivity under mAP(\%)(VOC12) on the HRSC2016 dataset}}
\label{table3_2}
\begin{threeparttable}
\resizebox*{\linewidth}{!}{
\setlength{\tabcolsep}{1.6mm}{
\renewcommand\arraystretch{1.2}{
\textcolor{black}{
\begin{tabular}{c|c|ccc}
\hline\hline
Methods                                                                          & $C_\theta$                          & mAP$_{50}$                          & mAP$_{85}$     & mAP$_{50:95}$   \\
\hline
\multirow{5}{*}{\begin{tabular}[c]{@{}c@{}}baseline\\+ DCL(binary)\end{tabular}} & 32                                  & 94.15                               & 13.97          & 53.28           \\
                                                                                 & 64                                  & 90.10                               & 7.95           & 46.80           \\
                                                                                 & 128                                 & 93.86                               & 18.06          & 56.23           \\
                                                                                 & 256                                 & 93.67                               & 15.34          & 54.6            \\
\cline{2-5}
                                                                                 & \multicolumn{1}{l|}{$\mu\pm\sigma$} & 93.84$\pm$1.91                      & 19.81$\pm$4.27 & 55.85$\pm$4.13  \\
\hline
\multirow{5}{*}{\begin{tabular}[c]{@{}c@{}}baseline\\+ DCL(gray)\end{tabular}}   & 32                                  & 97.02                               & 32.13          & 65.65           \\
                                                                                 & 64                                  & 97.44                               & 42.09          & 68.18           \\
                                                                                 & 128                                 & 97.26                               & 34.54          & 65.91           \\
                                                                                 & 256                                 & 97.28                               & 24.28          & 62.32           \\
\cline{2-5}
                                                                                 & \multicolumn{1}{l|}{$\mu\pm\sigma$} & 97.25$\pm$0.15                      & 33.26$\pm$6.35 & 65.52$\pm$2.09  \\
\hline
\multirow{4}{*}{\begin{tabular}[c]{@{}c@{}}baseline\\+ MGAR\end{tabular}}        & 5                                   & 97.62                               & 49.58          & 68.83           \\
                                                                                 & 4                                   & 97.19                               & 48.88          & 68.77           \\
                                                                                 & 3                                   & 97.46                               & 48.66          & 68.66           \\
\cline{2-5}
                                                                                 & \multicolumn{1}{l|}{$\mu\pm\sigma$} & \multicolumn{1}{c|}{\textbf{97.42$\pm$0.17}} & \textbf{49.04$\pm$0.39} & \textbf{68.75$\pm$0.07}  \\
\hline\hline
\end{tabular}
}
}}}
\begin{tablenotes}
\footnotesize
\item {\textcolor{black}{Note: $\mu$ represents Average, and $\sigma$ represents Standard Deviation}}
\end{tablenotes}
\end{threeparttable}
\end{table}

\subsection{Experimental Datasets}
\subsubsection{HRSC2016}
HRSC2016 \cite{liu2017hwigh} is an oriented ship detection dataset, including
two main scenes of nearshore and offshore, with 2976 ship targets. The image
size varies from 300$\times$300 pixels to 1500$\times$900 pixels. The dataset is divided into a training set with 436 images, a validation set with 181 images, and a testing set with 444 images.
\subsubsection{DOSR}
DOSR \cite{han2021fine} is a dataset for oriented ship recognition. The DOSR
dataset is mainly collected from Google Earth, including 1066 optical remote
sensing images and 6127 ship instances. The image size is ranged from 600 to
1300 pixels with resolutions of 0.5-2.5 m. The dataset contains rich scenes,
including nearshore scenes and offshore scenes. And the dataset contains 20
fine-grained classes of ships, which are Submarine (Sub), Tanker (Tan), Bulk Cargo
Vessel (BCV), Auxiliary Ship (Aux), Yacht (Yac), Military Ship (Mil), Barge (Bar),
Flat Traffic Ship (FTS), Deck Barge (DeB), Cruise (Cru), Container (Con),
Cargo (Car), Transport (Tra), Deck Ship (DeS), Floating Crane (Flo), Fishing
Boat(Fis), Tug, Communication Ship (Com), Multihull (Mul), and Speedboat (Spe). The
object distribution of this dataset belongs to the long-tail distribution.

\begin{figure*}[htbp]
        \centering
        \includegraphics[width=1.0\textwidth]{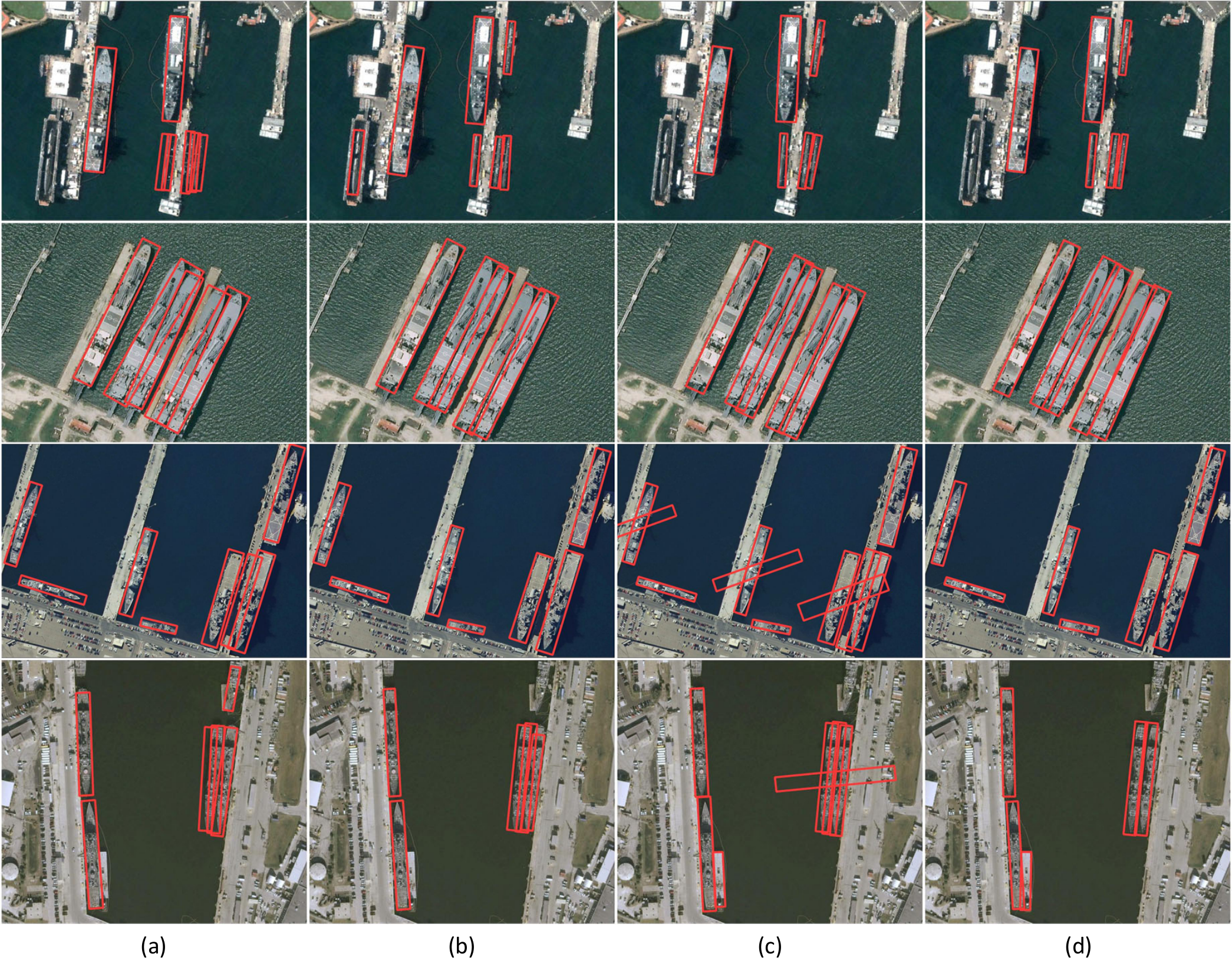}
        \caption{Comparison of the visualization results of four methods for complex scenarios on the HRSC2016 dataset. (a) The Regression-based method. (b) The CSL method. (c) THe DCL (gray) method. (d) The proposed MGAR method.}
        \label{fig4}
\end{figure*}

\begin{table}[htbp]
\centering
\caption{\textcolor{black}{Ablation studies for larger $C_\theta$ under mAP(\%)(VOC12) on the HRSC2016 dataset}}
\label{table3_3}
\resizebox*{\linewidth}{!}{
\setlength{\tabcolsep}{3.0mm}{
\renewcommand\arraystretch{1.1}{
\textcolor{black}{
\begin{tabular}{c|c|ccc}
\hline\hline
Methods                                                                    & $C_\theta$ & mAP$_{50}$ & mAP$_{85}$ & mAP$_{50:95}$  \\
\hline
\multirow{11}{*}{\begin{tabular}[c]{@{}c@{}}baseline\\+ MGAR\end{tabular}} & 6          & 96.95      & 47.64      & 68.57          \\
                                                                           & 9          & 97.32      & 46.45      & 68.12          \\
                                                                           & 10         & 97.39      & 43.41      & 66.45          \\
                                                                           & 12         & 97.79      & 38.99      & 65.82          \\
                                                                           & 15         & 97.39      & 18.87      & 58.16          \\
                                                                           & 18         & 97.72      & 22.10      & 61.14          \\
                                                                           & 20         & 97.50      & 22.43      & 61.69          \\
                                                                           & 30         & 97.93      & 29.64      & 65.46          \\
                                                                           & 45         & 97.75      & 39.88      & 67.29          \\
                                                                           & 60         & 97.68      & 37.48      & 66.90          \\
                                                                           & 90         & 97.71      & 37.15      & 66.48          \\
\hline\hline
\end{tabular}
}
}}}
\end{table}

\begin{figure}[htbp]
        \centering
        \includegraphics[width=0.9\linewidth]{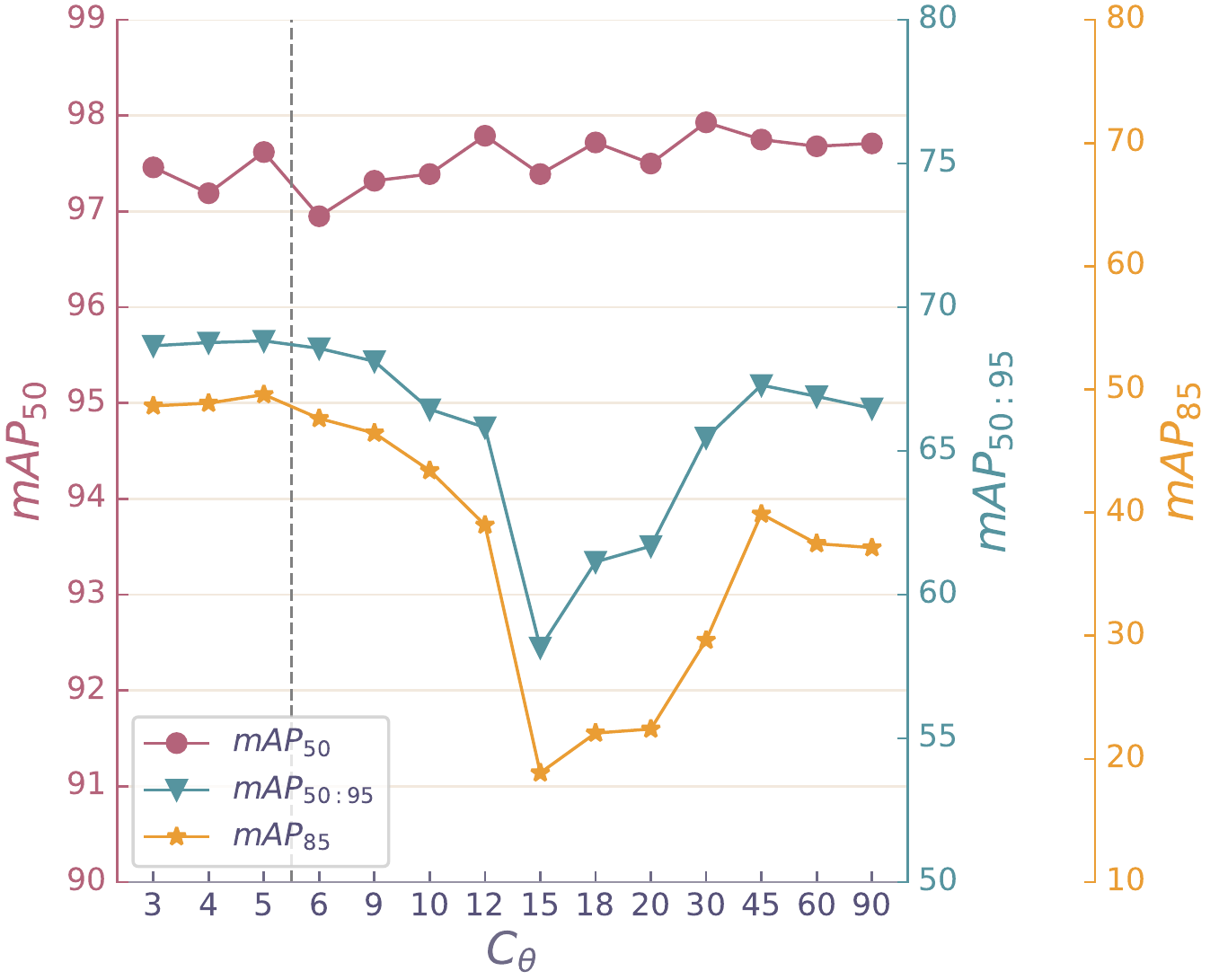}
        \caption{\textcolor{black}{Trend of the metrics mAP with respect to the hyperparameter $C_\theta$. The red curve represents mAP$_{50}$, the yellow curve represents mAP$_{85}$, the green curve represents mAP$_{50:95}$. }}
        \label{fig_z}
\end{figure}

\subsubsection{UCAS-AOD}
UCAS-AOD \cite{zhu2015orientation} contains 1510 aerial images, including two
categories of airplanes and cars, with 14596 instances. The sizes of images are mostly 659$\times$1280 pixels. According to the usual division manner, the UCAS-AOD dataset is divided into a training set with 755 images, a validation set with 302 images, and a testing set with 452 images.
\subsubsection{DIOR-R}
The DIOR-R \cite{cheng2022anchor} dataset is based on the DIOR dataset with
OBB-based annotations added. The dataset consists of 23463 images and 190288
object instances with 20 classes. The size of all images are 800$\times$800 pixels.
\subsubsection{DOTA}
The DOTA \cite{xia2018dota} dataset is a classic remote sensing scene rotating
object dataset, which contains 2806 huge remote sensing images ranging from 800
pixels to 4000 pixels. There are 15 classes in DOTA, including Plane (PL),
Baseball diamond (BD), Bridge (BR), Ground field track (GFT), Small vehicle
(SV), Large vehicle (SV), Ship (SH), Tennis court (TC), Basketball court (BC),
Storage tank (ST), Soccer-ball field (SBF), Roundabout (RA), Harbor (HA),
Swimming pool(SP), and Helicopter (HC).

\subsection{Evaluation Metrics and Implementation Details}
\subsubsection{Evaluation Metrics}
For AOOD methods, the mainstream evaluation metric is mean Average
Precision(mAP), which is consistent with HBB, and IoU is calculated by rotating
IoU with angles. We use two mAP calculation methods based on PASCAL VOC: VOC07 \cite{everingham2010pascal}
and VOC12 \cite{everingham2015pascal}. Usually, mAP is calculated according to an IoU threshold of 0.5. In
some cases where we need to compare higher accuracy, the results of mAP
calculations at IoU thresholds of 0.75 and 0.85 are used.
\textcolor{black}{We use the detected frames per second (FPS)} to evaluate the speed of detection.
\textcolor{black}{The FLOPs is used to evaluate the theoretical computational complexity of a model.}
And the number of parameters is used to evaluate the model size.

\begin{table}[htbp]
        \centering
        \caption{Comparison of different regression fitting functions on MGAR under mAP(\%)(VOC07) }
        \label{table4}
                \setlength{\tabcolsep}{0.8mm}{
                        \renewcommand\arraystretch{1.4}{
                                \begin{tabular}{c|c|c|cccc}
                                        \hline\hline
                                        Methods                          & $C_{\theta}$ & Function & mAP$_{50}$\textit{} & mAP$_{75}$       & mAP$_{85}$       & mAP$_{50:95}$    \\
                                        \hline
                                        \multirow{4}{*}{baseline+MGAR~} & \multirow{4}{*}{3} & $Linear$   & 97.36               & \textbf{82.74} & 47.29            & \textbf{68.88} \\
                                                                        &                        & $Sigmoid$  & 97.18               & 79.76            & 43.59            & 67.14            \\
                                                                        &                        & $Square$   & \textbf{97.46}    & 82.05            & \textbf{48.66} & 68.66            \\
                                                                        &                        & $Exp$      & 96.78               & 74.99            & 38.74            & 64.56            \\
                                        \hline\hline
                                \end{tabular}
}}
\end{table}

\begin{table}[htbp]
        \centering
        \caption{Comparison of different regression loss functions on MGAR under mAP(\%)(VOC07)}
        \label{table5}
                \setlength{\tabcolsep}{1.2mm}{
                        \renewcommand\arraystretch{1.4}{
                                \begin{tabular}{c|c|c|cccc}
                                        \hline\hline
                                        Methods                          & $C_{\theta}$               & Loss & mAP$_{50}$       & mAP$_{75}$       & mAP$_{85}$       & mAP$_{50:95}$    \\
                                        \hline
                                        \multirow{2}{*}{baseline+MGAR~} & \multirow{2}{*}{3} & MSE  & 93.68            & 67.74            & 25.26            & 58.68            \\
                                                                        &                        & IFL & \textbf{97.46} & \textbf{82.05} & \textbf{48.66} & \textbf{68.66} \\

                                        \hline\hline
                                \end{tabular}}}
\end{table}

\begin{table}[htbp]
        \centering
        \caption{Comparative performance (mAP(\%)(VOC07) and mAP(\%)(VOC12)) of different methods on the HRSC2016 dataset}
        \label{table6}
                \setlength{\tabcolsep}{1.8mm}{
                        \renewcommand\arraystretch{1.4}{
                                \begin{tabular}{l|c|cc}
                                        \hline\hline
                                        Methods                                    & Backbone  & mAP(07) & mAP(12) \\
                                        \hline
                                        R$^2$CNN \cite{jiang2017r2cnn}           & ResNet101 \cite{he2016deep} & 73.07   & 79.73   \\
                                        RoI-Transformer \cite{ding2019learning}  & ResNet101 & 86.20   & -       \\
                                        Gliding Vertex \cite{xu2020gliding}      & ResNet101 & 88.20   & -       \\
                                        BBAVectors  \cite{yi2021oriented}        & ResNet101 & 88.6    & -       \\
                                        CenterMap OBB \cite{wang2020learning}    & ResNet50  & -       & 92.8    \\
                                        RetinaNet-R \cite{yang2021r3det}         & ResNet101 & 89.18   & 95.21   \\
                                        R$^3$Det \cite{yang2021r3det}            & ResNet101 & 89.26   & 96.01   \\
                                        R$^3$Det-DCL \cite{yang2021dense}        & ResNet101 & 89.46   & 96.41   \\
                                        S$^2$ANet \cite{han2021align}            & ResNet101 & 90.17   & 95.01   \\
                                        Oriented RepPoints \cite{li2022oriented} & ResNet50  & \textbf{90.40}   & 97.26   \\
                                        \hline
                                        MGAR($C_{\theta}=3$)             & DarkNet53 & 90.32   & \textbf{97.46}  \\
                                        \hline\hline
                                \end{tabular}}}
\end{table}

\begin{table}[htbp]
        \centering
        \caption{Comparative high-accuracy performance (mAP(\%)(VOC07)) of different methods on the HRSC2016 dataset}
        \label{table7}
                \setlength{\tabcolsep}{4mm}{
                        \renewcommand\arraystretch{1.4}{
                \begin{tabular}{l|ccc}
                        \hline\hline
                        Methods            & mAP$_{50}$       & mAP$_{75}$       & mAP$_{85}$       \\
                        \hline
                        RetinaNet(GWD) \cite{yang2021rethinking}     & 85.56            & 60.31            & 17.14            \\
                        RetinaNet(KLD) \cite{https://doi.org/10.48550/arxiv.2106.01883}    & 87.45            & 72.39            & 27.68            \\
                        R3Det(GWD) \cite{yang2021rethinking}        & 89.43            & 68.88            & 15.02            \\
                        R3Det(KLD) \cite{https://doi.org/10.48550/arxiv.2106.01883}        & 89.97            & 77.38            & 25.12            \\
                        \hline
                        MGAR($C_{\theta}=5$) & \textbf{90.32} & \textbf{79.19} & \textbf{46.25} \\
                        \hline\hline
                \end{tabular}
        }}
\end{table}

\begin{table*}[htbp]
        \normalsize
        \centering
        \caption{Comparative performance (AP(\%) and mAP(\%)(VOC07)) of different methods on the DOSR dataset. $^*$ means form \cite{han2021fine}}
        \label{table8}
        \begin{threeparttable}
        \resizebox{\linewidth}{!}{
                \setlength{\tabcolsep}{0.5mm}{
                        \renewcommand\arraystretch{1.3}{
                                \begin{tabular}{l|ccccccccccccccccccccc|c}
                                        \hline\hline
                                        Methods                            & BCV.             & Fis.             & DeB.             & Yac.             & FTS.             & Mul.             & Tug              & Com.             & Spe.             & Car.             & Cru.             & Flo.            & Tan.             & DeS.             & Sub.             & Con.           & Bar.             & Tra.             & Aux.             & Mil.             & mAP              &\begin{tabular}[c]{@{}c@{}}Speed\\(FPS)\end{tabular}         \\
                                        \hline
                                        FR-FPN-O \cite{han2021fine}                & 37.14            & 8.79             & 7.98             & 47.30            & 25.59            & 48.20            & 48.11            & 50.80            & 28.54            & 80.57            & 49.81            & 8.64            & 65.40            & 15.54            & 10.03            & 56.69          & 12.21            & 69.35            & 10.78            & 32.18            & 35.68            & 4.92            \\
                                        R$^2$CNN \cite{jiang2017r2cnn}$^*$  & 56.47            & 36.86            & 38.84            & 57.14            & 26.19            & 54.55            & 58.98            & 32.03            & 39.27            & 76.70            & 52.99            & 5.85            & 50.84            & 42.41            & 13.64            & 75.57          & 46.08            & 66.85            & 43.54            & 36.85            & 45.58            & 2.85            \\
                                        RRPN \cite{ma2018arbitrary}$^*$      & 62.89            & 42.80            & 33.15            & 47.15            & 43.66            & 52.50            & 74.57            & 27.27            & 31.19            & 83.30            & 57.52            & 33.10           & 64.89            & 20.55            & 54.55            & 78.61          & \textbf{52.95} & 78.26            & 36.06            & 36.63            & 50.58            & 2.71            \\
                                        SCRDet \cite{yang2019scrdet}$^*$     & 65.21            & 54.26            & 44.62            & 53.33            & 49.46            & 38.31            & 66.67            & 40.74            & 32.19            & 86.66            & 61.50            & 21.95           & \textbf{87.68} & 32.84            & 54.55            & \textbf{80.58} & 49.91            & 73.68            & 42.78            & 49.46            & 54.29            & 3.84           \\
                                        RetinaNet-O \cite{yang2021r3det}$^*$ & 49.52            & 26.24            & 36.86            & 50.63            & 27.16            & 35.00            & 70.45            & 49.74            & 33.68            & 68.70            & 35.87            & 11.93           & 38.11            & 2.04             & 0.00             & 55.99          & 12.16            & 66.27            & 29.66            & 34.06            & 36.70            & 3.72            \\
                                        R$^3$Det \cite{yang2021r3det}$^*$    & 65.60            & 44.78            & 28.49            & 64.21            & 49.14            & 53.03            & 68.16            & 38.83            & 35.97            & 84.24            & 63.36            & 34.35           & 71.63            & 26.70            & 57.61            & 74.77          & 42.70            & 74.03            & 27.19            & 48.48            & 52.66            & 4.89           \\
                                        SCRDet++ \cite{yang2022scrdet++}$^*$ & 61.22            & 44.45            & 36.06            & 67.58            & 62.80            & 61.62            & \textbf{81.09} & \textbf{66.09} & 63.51            & 76.28            & 66.14            & 13.91           & 77.01            & 58.98            & 32.73            & 70.00          & 18.75            & 76.22            & \textbf{47.27} & 44.22            & 56.33            & 2.85            \\
                                        RSDet \cite{qian2021learning}$^*$   & 55.27            & 10.76            & 22.84            & 59.78            & 52.56            & 47.30            & 58.78            & 63.64            & 57.72            & 75.32            & 39.60            & 16.94           & 33.13            & 14.23            & 45.69            & 68.82          & 26.19            & 75.30            & 33.96            & 45.04            & 45.15            & 2.77            \\
                                        ReDet \cite{han2021redet}$^*$       & 64.28            & 43.17            & 27.87            & \textbf{75.65} & 65.06            & 41.59            & 80.10            & 37.67            & \textbf{67.95} & 85.06            & 62.40            & 37.54           & 83.44            & 36.58            & 34.85            & 76.64          & 44.31            & 86.03            & 38.04            & 57.48            & 57.32            & 12.00            \\
                                        EIRNet \cite{han2021fine}        & 67.52            & \textbf{55.65} & 48.59            & 68.53            & \textbf{70.48} & 57.36            & 75.52            & 58.26            & 42.53            & \textbf{87.87} & \textbf{67.00} & 55.30           & 74.90            & 56.65            & \textbf{59.60} & 78.63          & 30.46            & 78.34            & 45.98            & 48.59            & 61.39            & 3.68            \\
                                        \hline
                                        MGAR$^{\dag}$     & \textbf{68.12} & 41.68            & \textbf{51.98} & 71.78            & 63.64            & \textbf{81.82} & 78.96            & 63.64            & 66.22            & 86.41            & 59.01            & \textbf{55.31} & 82.39            & \textbf{63.27} & 21.82            & 75.55          & 25.97            & \textbf{88.93} & 40.33            & \textbf{57.65} & \textbf{62.22} & \textbf{26.17} \\
                                        \hline\hline
                                \end{tabular}}}}
                        \begin{tablenotes}
                        \footnotesize
                        \item {Note: The unit FPS means Frames Per Second. To maintain consistency with other methods, we test speed on the same equipment: NVIDIA GeForce \\ GTX 1080Ti. $^{\dag}$ means $C_{\theta}=3$.}

                        \end{tablenotes}
                \end{threeparttable}
\end{table*}

\begin{table}[htbp]
        \centering
        \caption{Comparative performance (mAP(VOC07)) of different methods on the UCAS-AOD dataset}
        \label{table9}
                \setlength{\tabcolsep}{3.2mm}{
                        \renewcommand\arraystretch{1.3}{
                                \begin{tabular}{l|ccc}
                                        \hline\hline
                                        Methods                               & Car              & Airplane         & mAP(07)          \\
                                        \hline
                                        YOLOv3-R \cite{redmon2018yolov3}    & 74.63            & 89.52            & 82.08            \\
                                        RetinaNet-R \cite{yang2021r3det}    & 84.63            & 90.51            & 87.57            \\
                                        Faster R-CNN-R \cite{ren2015faster} & 86.87            & 89.86            & 88.36            \\
                                        RoI-Transformer \cite{ding2019learning}  & 87.99            & 89.90            & 88.95            \\
                                        DAL \cite{ming2012dynamic}          & 89.25            & 90.49            & 89.87            \\
                                        S$^2$ANet \cite{han2021align}       & \textbf{89.56} & 90.42            & 89.99            \\
                                        \hline
                                        MGAR($C_{\theta}=5$)        & 89.40            & \textbf{90.63} & \textbf{90.01} \\
                                        \hline\hline
                                \end{tabular}}}
\end{table}

\begin{table}[htbp]
        \centering
        \caption{Comparative performance (mAP(VOC07)) of different methods on the DIOR-R dataset}
        \label{table10}
        \begin{threeparttable}
                \setlength{\tabcolsep}{5.0mm}{
                        \renewcommand\arraystretch{1.3}{
                                \begin{tabular}{l|c|c}
                                        \hline\hline
                                        Methods                                   & Backbone   & mAP(07)          \\
                                        \hline
                                        RetinaNet-O \cite{lin2017focal}          & ResNet-50  & 57.55            \\
                                        Faster RCNN-O \cite{ren2015faster}       & ResNet-50  & 59.54            \\
                                        Gliding Vertex \cite{xu2020gliding}      & ResNet-50  & 60.06            \\
                                        RoI-Transformer \cite{ding2019learning}       & ResNet-50  & 63.87            \\
                                        AOPG \cite{cheng2022anchor}              & ResNet-50  & 64.41            \\
                                        Oriented RepPoints \cite{li2022oriented} & ResNet-50  & 66.71            \\
                                        \hline
                                        MGAR($C_{\theta}=5$)             & Darknet-53 & \textbf{66.89} \\
                                        \hline\hline
                                \end{tabular}}}
                \end{threeparttable}
\end{table}

\subsubsection{Implementation Details}
All datasets are trained with training and validation sets.
The input image size is 800$\times$800 pixels for the HRSC2016, UCAS-AOD, and DIOR-R datasets.
The network input size for the DOSR dataset is set to 1024$\times$1024 pixels because of its large average image size and dense scenes.
For the DOTA dataset, because the original image size is large, the original image
is cropped by 800$\times$800 pixels with a stride of 200 pixels.
In addition, the zoom scale is [0.5, 1.0, 1.5] when cropping, and images with the size of 896$\times$896 pixels are used as network input for both training and testing.
The data augmentation strategy is adopted to alleviate the over-fitting problem, including horizontal and vertical flipping, random color transformation of \textcolor{black}{HSV(hue, saturation, value) color gamut}, mixup \cite{zhang2017mixup}, and random rotation in three directions of $90^{\circ}$, $180^{\circ}$, and $270^{\circ}$.
For batch size, it is 16 when the input image size is 800$\times$800 pixels and 8 when the input image size is 1024$\times$1024 pixels. \textcolor{black}{The stochastic gradient descent (SGD) optimizer} with a momentum of 0.9 and a weight decay of $1\times10^{-5}$ is used during training.
The initial learning rate is set to $1\times10^{-4}$, and the final learning rate is $1\times10^{-5}$.

\subsection{Ablation Study}
In the case of the same angle deviation, objects with larger aspect ratios
cause greater deviation in calculating IoU, which further affects the mAP result.
The HRSC2016 dataset contains many ship objects with larger aspect ratios,
which helps to compare the fitting accuracy of the OBB. Therefore, the
HRSC2016 dataset is chosen to conduct ablation experiments,
comparing Regression, CSL, DCL, and the proposed MGAR.
\textcolor{black}{We take the regression-based method as the baseline and implement the other three methods on the basis of it.
These four methods are consistent with the experimental settings except for their hyperparameters.
For the CSL, the optimal window size of the Gaussian function is 6.
DCL contains two types of encoding: binary encoding and gray encoding. When the hyperparameter $C_\theta \in [32, 64, 128, 256]$, DCL has the best performance. For the proposed MGAR, we choose $C_\theta \in [3, 4, 5]$ for comparison.
}

\begin{table*}[hbtp]
\centering
\caption{Comparative performance (mAP(\%) and Speed) of different methods on the DOTA dataset}
\label{table11}
\begin{threeparttable}
\resizebox{0.97\linewidth}{!}{
\setlength{\tabcolsep}{0.5mm}{
\renewcommand\arraystretch{1.3}{
\textcolor{black}{
\begin{tabular}{l|c|c|ccccccccccccccc|c|c}
\hline\hline
Methods                                                                       & Stage  & Backbone & PL             & BD             & BR             & GTF            & SV             & LV             & SH             & TC             & BC             & ST             & SBF            & RA             & HA             & SP             & HC             & mAP            & \begin{tabular}[c]{@{}c@{}}Speed\\(FPS)\end{tabular}  \\
\hline
Faster RCNN-O \cite{xia2018dota}                                                                & Two    & R-50     & 88.44          & 73.06          & 44.86          & 59.09          & 73.25          & 71.49          & 77.11          & 90.84          & 78.94          & 83.90          & 48.59          & 62.95          & 62.18          & 64.91          & 56.18          & 69.05          & 14.9                                                  \\
ROI-Transformer \cite{ding2019learning}                                                               & Two    & R-101    & 88.53          & 77.91          & 37.63          & 74.08          & 66.53          & 62.97          & 66.57          & 90.50          & 79.46          & 76.75          & 59.04          & 56.73          & 62.54          & 61.29          & 55.56          & 67.74          & 7.80                                                  \\
SCRDet \cite{yang2019scrdet}                                                                       & Two    & R-101    & 89.98          & 80.65          & 52.09          & 68.36          & 68.36          & 60.32          & 72.41          & 90.85          & 87.94          & 86.86          & 65.02          & 66.68          & 66.25          & 68.24          & 65.21          & 72.61          & 9.51                                                  \\
Gliding Vertex \cite{xu2020gliding}                                                               & Two    & R-101    & 89.64          & 85.00          & 52.26          & 77.34          & 73.01          & 73.14          & 86.82          & 90.74          & 79.02          & 86.81          & 59.55          & \textbf{70.91} & 72.94          & 70.86          & 57.32          & 75.02          & 13.10                                                 \\
ReDet \cite{han2021redet}                                                                        & Two    & ReR-50   & 88.79          & 82.64          & 53.97          & 74.00          & 78.13          & 84.06          & 88.04          & 90.89          & 87.78          & 85.75          & 61.76          & 60.39          & 75.96          & 68.07          & 63.59          & 76.25          & -                                                     \\
Oriented R-CNN \cite{xie2021oriented}                                                              & Two    & R-50     & 89.84          & 95.43          & \textbf{61.09} & 79.82          & 79.71          & \textbf{85.35} & 88.82          & 90.88          & 86.68          & 87.73          & 72.21          & 70.80          & 82.42          & 78.18          & \textbf{74.11} & \textbf{80.87} & 8.10                                                  \\
RSDet \cite{qian2021learning}                                                                        & Two    & R-101    & 89.80          & 82.90          & 48.60          & 65.20          & 69.50          & 70.10          & 70.20          & 90.50          & 85.60          & 83.40          & 62.50          & 63.90          & 65.60          & 67.20          & 68.00          & 72.20          & -                                                     \\
\hline
R$^3$Det \cite{yang2021r3det}                                                                     & Refine & R-152    & 89.80          & 83.77          & 48.11          & 66.77          & 78.76          & 83.27          & 87.84          & 90.82          & 85.38          & 85.51          & 65.67          & 62.68          & 67.53          & 78.56          & 72.62          & 76.47          & 10.53                                                 \\
\hline
\multirow{2}{*}{S$^2$ANet \cite{han2021align}}                                                    & Refine & R-50     & 89.07          & 82.22          & 53.63          & 69.88          & 80.94          & 82.12          & 88.72          & 90.73          & 83.77          & 86.92          & 63.78          & 67.86          & 76.51          & 73.03          & 56.60          & 76.38          & 17.60                                                 \\
                                                                              & Refine & R-101    & 88.89          & 83.60          & 57.74          & \textbf{81.95} & 79.94          & 83.19          & \textbf{89.11} & 90.78          & 84.87          & 87.81          & 70.30          & 68.25          & \textbf{78.30} & 77.01          & 69.58          & 79.42          & 13.79                                                 \\
\hline
\multirow{3}{*}{\begin{tabular}[l]{@{}l@{}}Oriented\\Reppionts \cite{li2022oriented}\end{tabular}} & Refine & R-50     & 87.02          & 83.17          & 54.13          & 71.16          & 80.18          & 78.40          & 87.28          & \textbf{90.90} & 85.97          & 86.25          & 59.90          & 70.49          & 73.33          & 72.27          & 58.97          & 75.97          & 16.10                                                 \\
                                                                              & Refine & R-101    & 88.86          & \textbf{88.86} & 55.27          & 76.92          & 74.27          & 82.10          & 87.52          & \textbf{90.90} & 85.56          & 85.33          & 65.51          & 66.82          & 74.36          & 70.15          & 57.28          & 76.28          & 14.23                                                 \\
                                                                              & Refine & Swin-T   & 88.72          & 80.56          & 55.69          & 75.07          & \textbf{81.84} & 82.40          & 87.97          & 90.80          & 84.33          & 87.64          & 62.80          & 67.91          & 77.69          & \textbf{82.94} & 65.46          & 78.12          & -                                                     \\
\hline
DRN \cite{pan2020dynamic}                                                                          & One    & H-104    & 89.71          & 82.34          & 47.22          & 64.10          & 76.22          & 74.43          & 85.84          & 90.57          & 86.18          & 84.89          & 57.65          & 61.93          & 69.30          & 69.63          & 58.48          & 73.23          & -                                                     \\
RIDet \cite{ming2021optimization}                                                                        & One    & R-101    & 88.93          & 78.45          & 46.87          & 72.63          & 77.63          & 80.68          & 88.18          & 90.55          & 81.33          & 83.61          & 64.85          & 63.72          & 73.09          & 73.13          & 56.87          & 74.70          & 13.36                                                 \\
PolarDet \cite{zhao2021polardet}                                                                     & One    & R-101    & 89.65          & 87.07          & 48.14          & 70.97          & 78.53          & 80.;34         & 87.45          & 90.76          & 85.63          & 86.87          & 61.64          & 70.32          & 71.92          & 73.09          & 67.15          & 76.64          & 25.00                                                 \\
GGHL \cite{9709203}                                                                         & One    & D-53     & 89.74          & 85.63          & 44.50          & 77.48          & 76.72          & 80.45          & 86.16          & 90.83          & \textbf{88.18} & 86.25          & 67.07          & 69.40          & 73.38          & 68.45          & 70.14          & 76.95          & 42.30                                                 \\
GWD \cite{yang2021rethinking}                                                                          & One    & R-152    & 86.96          & 83.88          & 54.36          & 77.53          & 74.41          & 68.48          & 80.34          & 86.62          & 83.41          & 85.55          & \textbf{73.47} & 67.77          & 72.57          & 75.76          & 73.40          & 76.30          & 13.86                                                 \\
KFIoU \cite{yang2022kfiou}                                                                        & One    & R-152    & 89.46          & 85.72          & 54.94          & 80.37          & 77.16          & 69.23          & 80.90          & 90.79          & 87.79          & 86.13          & 73.32          & 68.11          & 75.23          & 71.61          & 69.49          & 77.35          & 13.79                                                 \\
\hline
CSL \cite{yang2020arbitrars}                                                                          & Two    & R-152    & \textbf{90.25} & 85.53          & 54.64          & 75.31          & 70.44          & 73.51          & 77.62          & 90.84          & 86.15          & 86.69          & 69.60          & 68.04          & 73.83          & 71.10          & 68.93          & 76.17          & 8.89                                                  \\
DCL \cite{yang2021dense}                                                                          & Refine & R-152    & 89.26          & 83.60          & 53.54          & 72.76          & 79.04          & 82.56          & 87.31          & 90.67          & 86.59          & 86.98          & 67.49          & 66.88          & 73.29          & 70.56          & 69.99          & 77.37          & 10.39                                                 \\
\hline
MGAR$^*(C_\theta=3)$                                                                     & One    & D-53     & 89.84          & 85.75          & 51.59          & 77.00          & 76.38          & 74.81          & 86.40          & 90.73          & 87.70          & 87.48          & 63.25          & 69.70          & 75.79          & 80.88          & 71.07          & 77.85          & \textbf{59.17}                                                 \\
MGAR$^\dag(C_\theta=5)$                                                                   & One    & D-53     & 89.81          & 85.22          & 52.51          & 77.52          & 77.63          & 76.19          & 87.20          & 90.84          & 87.93          & \textbf{88.01} & 66.25          & 67.88          & 76.24          & 78.53          & 72.51          & 78.29          & 58.14                                        \\
\hline\hline
\end{tabular}
}
}}}

\begin{tablenotes}
\footnotesize
\item{Note:
\textcolor{black}{The backbone networks R-50, R-101, R-152, ReR-50, H-104, D-53, Swin-T, represent the ResNet50 \cite{he2016deep}, ResNet101 \cite{he2016deep}, ResNet152 \cite{he2016deep}, ReResNet50 \cite{han2021redet}, Hourglass104 \cite{zhou2019objects}, DarkNet53 \cite{redmon2018yolov3}, and Swin Transformer Tiny \cite{liu2021swin}, respectively. "One", "Two", "Refine" represent the one-stage method, two-stage method and refine-stage method, respectively.
Speed is the test result on NVIDIA GeForce RTX 3090. The speed (average of 10 tests) only includes the network inference speed without post-processing (batch size = 1).
When testing other methods, their open source codes are used.
The speed of some methods could not be tested due to the available codes, which is indicated by "-".
$^*$ means the input size of image for network is 800$\times$800 pixels. $^{\dag}$ means the input size of image for network is 896$\times$896 pixels.}
}
\end{tablenotes}
\end{threeparttable}
\end{table*}

\begin{figure*}
        \centering
        \includegraphics[width=1.0\textwidth]{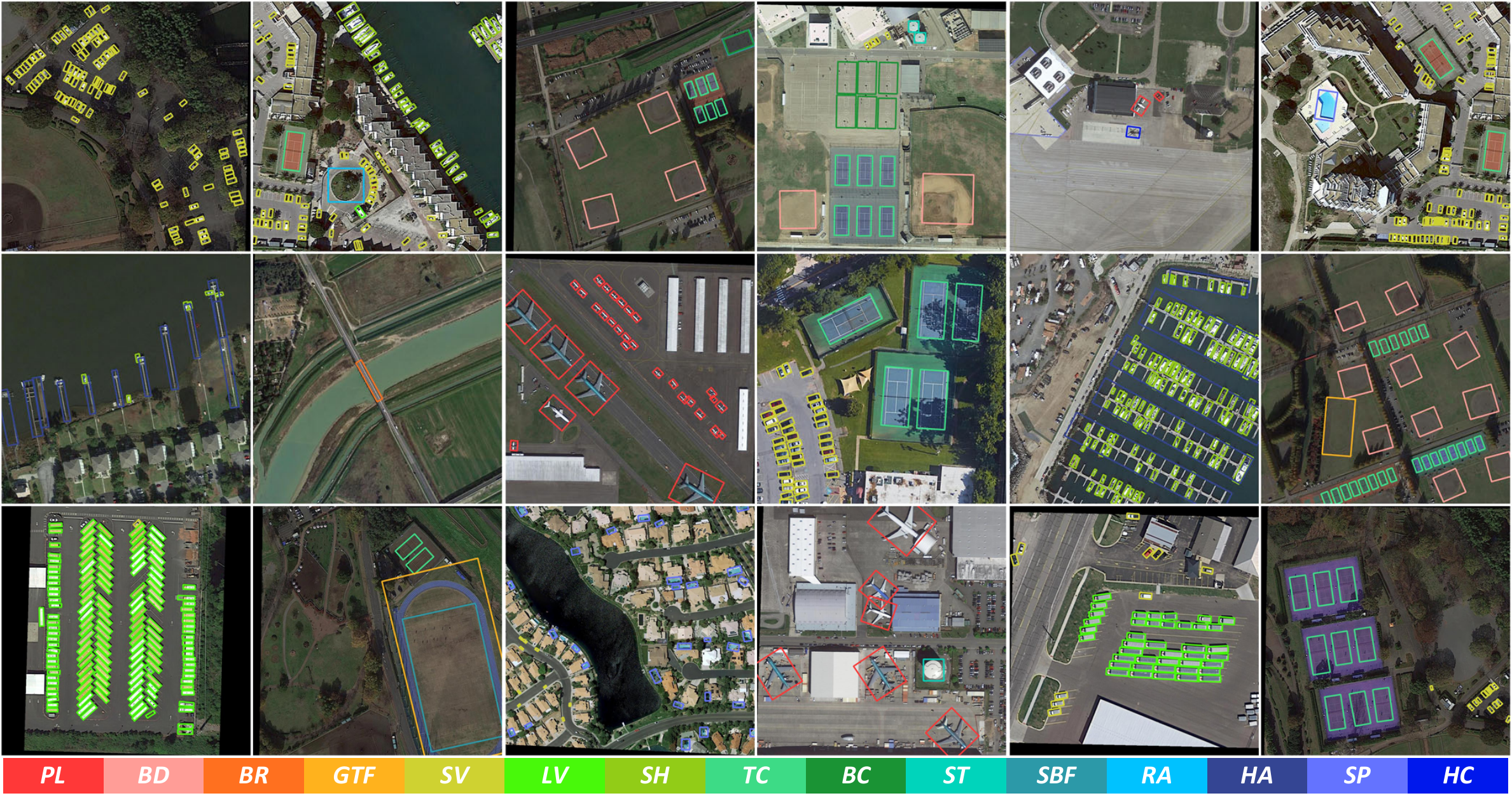}
        \caption{Visualization results of baseline+MGAR method on the DOTA dataset.}
        \label{fig5}
\end{figure*}

\subsubsection{Comparison of Four Methods}
\textcolor{black}{
The best results of these four methods are listed in Table \ref{table3_1}.
Three accuracy metrics are adopted to evaluate the performance of these four methods.
The mAP$_{50}$ is a commonly used base metric, the mAP$_{85}$ evaluates the accuracy with more stringent criteria, and the mAP$_{50:95}$ aims to evaluate the overall performance of the methods in a more comprehensive way.
We compare the differences in model performance before and after the addition of the SPP module on a regression-based method, and from the results, SPP helps to improve accuracy.
Compare to other methods, the proposed method MGAR performs best, with the mAP$_{50}$, mAP$_{85}$, and mAP$_{50:95}$ reaching 62\%, 49.58\%, and 68.83\%, respectively.
In particular, the mAP$_{85}$ is higher than baseline, CSL, DCL(binary) and DCL (gray) respectively than 34.61\%, 5.86\%, 31.52\%, 7.49\%.
which shows that FAR can get fine-grained angle information more accurately.
In addition, for the speed metric,
because CSL, DCL, and MGAR all have additional decoding operations, which increase the time overhead in the Non-maximum Suppression (NMS) stage, the time listed in Table \ref{table3_1} includes network inference time wih post-processing.
It is observed that our proposed method is the fastest, reaching 56.21 FPS, which is beneficial for lightweight deployment.
In summary, the proposed MGAR has an advantage over other methods in terms of speed and accuracy.
}

\textcolor{black}{
We also analyze the sensitivity of hyperparameters $C_{\theta}$ in different methods.
Table \ref{table3_2} lists the results of DCL (gray), DCL (binary), and the proposed MGAR in the respective optimal hyperparameter range.
The results show that different values of $C_{\theta}$ have large impact on the DCL.
It is seen that the proposed MGAR has the best average metric results with the smallest standard deviation, which indicates that the hyperparameters of the proposed MGAR are insensitive.
In contrast, the two DCL methods are less stable and more sensitive to $C_{\theta}$, which implicitly increases the cost of training time on different datasets.
}

MGAR develops a method based on coarse-grained classification and fine-grained
regression, which effectively combines the flexibility of classification and
the accuracy of regression, reducing the learning difficulty of fine-grained
classification and large-scale regression. Fig. \ref{fig4} visualizes the
detection results of four state-of-the-art methods in complex scenes, and it can be observed
that the proposed MGAR has better performance in detail.

\subsubsection{Impact of Coarse-Grained Angle Classification (CAC)}
\textcolor{black}{
Hyperparameter $C_\theta$ mainly affects the classification granularity of the angle.
For the proposed MGAR, the value of $C_\theta$ theoretically determines the thickness of the detection head,
which in turn affects the speed of the model.
To more fully analyze the impact of $C_\theta$ on the MGAR method, we analyze the results for $C_\theta$ at larger values.
Considering the floating-point error caused by $C_\theta$ not being divisible by $180^{\circ}$,
we discuss the cases of $C_\theta = [6, 9, 10, 12, 15, 18, 20, 30, 45, 60, 90]$.
Table \ref{table3_3} presents the experimental results, and the influence of $C_\theta$ on the three accuracy metrics is shown in Fig. \ref{fig_z}.
It is observed that as $C_\theta$ increases, there is a partial increase in the mAP$_{50}$, while the mAP$_{85}$ and mAP$_{50:95}$ decrease to some extent.
This is mainly because the increase of angle categories increases the difficulty of network classification.
To alleviate this problem, the CSL smooths the One-hot label by introducing the Gaussian window function.
However, this method introduces an additional window size hyperparameter, further increasing the parameter selection and invisible training costs.
Therefore, to balance accuracy, speed, and parameter stability, the selection range of $C_\theta$ is restricted to $[3, 4, 5]$ in the proposed MGAR.
}

\subsubsection{Impact of Fine-Grained Angle Regression (FAR)}
For angle regression, the proposed MGAR has a clear regression range of \textcolor{black}{$[0^{\circ},
        \omega)$}.
For the predict values of the angle regression, $\theta_{reg.}^{'}=Function(t_{\theta_{reg.}}^{'})$ is used to represent the fitting function of the network predict value $t_{\theta_{reg.}}^{'}$.
There are many fitting functions to
choose from, such as $Linear$, $Sigmoid$, $Square$, and $Exp$. $Sigmoid$ function can conform to
the angle regression range, and other functions theoretically have the
regression range of $[0, +\infty)$. Table \ref{table4} lists the performance of
different fitting functions. $Linear$ and $Square$ functions perform better than
$Sigmoid$ and $Exp$ functions, and $Square$ function is more smoother during training than $Linear$ function.
In general, the Square function performs best, so we finally choose it to fit the value of angle regression.
For the loss function of angle regression, we compare Mean-Squared (MSE) Loss  and IFL. The results are listed in Table \ref{table5}. The performance of IFL is better than that of MSE.

\subsection{Comparisons with State-of-the-Arts}
\subsubsection{Results on the HRSC2016 Dataset}
We compare other AOOD methods on the HRSC2016 dataset. The experimental results are shown in Table \ref{table6}.
\textcolor{black}{For convenience, we use "MGAR" to represent "baseline+MGAR" in the following tables.}
The results indicate that the proposed MGAR has achieved high performance, reaching 90.32\% and 97.46\% under the metrics mAP(07) and mAP(12), respectively. We also compare the results of different methods under higher detection performance, as listed in Table \ref{table7}. The proposed method achieves the best results of 79.19\% and 46.25\% under the two more difficult evaluation metrics of mAP$_{75}$ and mAP$_{85}$, respectively, which are higher than the Kullback-Leibler Divergence (KLD) \cite{https://doi.org/10.48550/arxiv.2106.01883} and GWD methods.

\subsubsection{Results on the DOSR Dataset}
To further verify the performance of the proposed method, we select a fine-grained long-tailed ship dataset DOSR with more data samples, more complex scenes, and more small objects. This dataset also presents a classification challenge to the AOOD methods. In Table \ref{table8}, we report the AP of each type of the ship target. The proposed method is better than other methods, achieving the best result of 62.22\% mAP and the fastest speed 26.17 FPS.

\subsubsection{Results on the UCAS-AOD Dataset}
The UCAS-AOD dataset contains densely arranged small-sized cars and aircraft at
different scales, which helps compare the detection effects of methods on
small objects. As shown in Table \ref{table9}, The proposed method can
achieve a performance of 90.01\% mAP, with excellent performance.

\subsubsection{Results on the DIOR-R Dataset}
DIOR-R contains 20 classes of objects in aviation scenarios. Table \ref{table10}
lists the experimental results on DIOR-R. Our method achieves the best
performance of 66.89\% mAP.

\subsubsection{Results on the DOTA Dataset}
We utilize the DOTA dataset to evaluate the performance of the proposed method on large datasets. \textcolor{black}{We compare the proposed MGAR with other two-stage methods, refine-stage methods, and single-stage methods.} We report the detailed results in Table \ref{table11}. Our method obtains 78.29\% mAP under the input size of 896$\times$896 pixels, \textcolor{black}{which is better than most of the two-stage methods and the refine-stage methods. At the same time, the proposed MGAR achieved best results in the single-stage methods.} It is worth noting that the proposed MGAR performs best in terms of speed, which is the fastest among all comparison methods. We also visualize some detection results on the DOTA testing set, as shown in Fig. \ref{fig5}. The proposed MGAR has excellent detection performance.

\begin{figure}[htbp]
        \centering
        \includegraphics[width=0.9\linewidth]{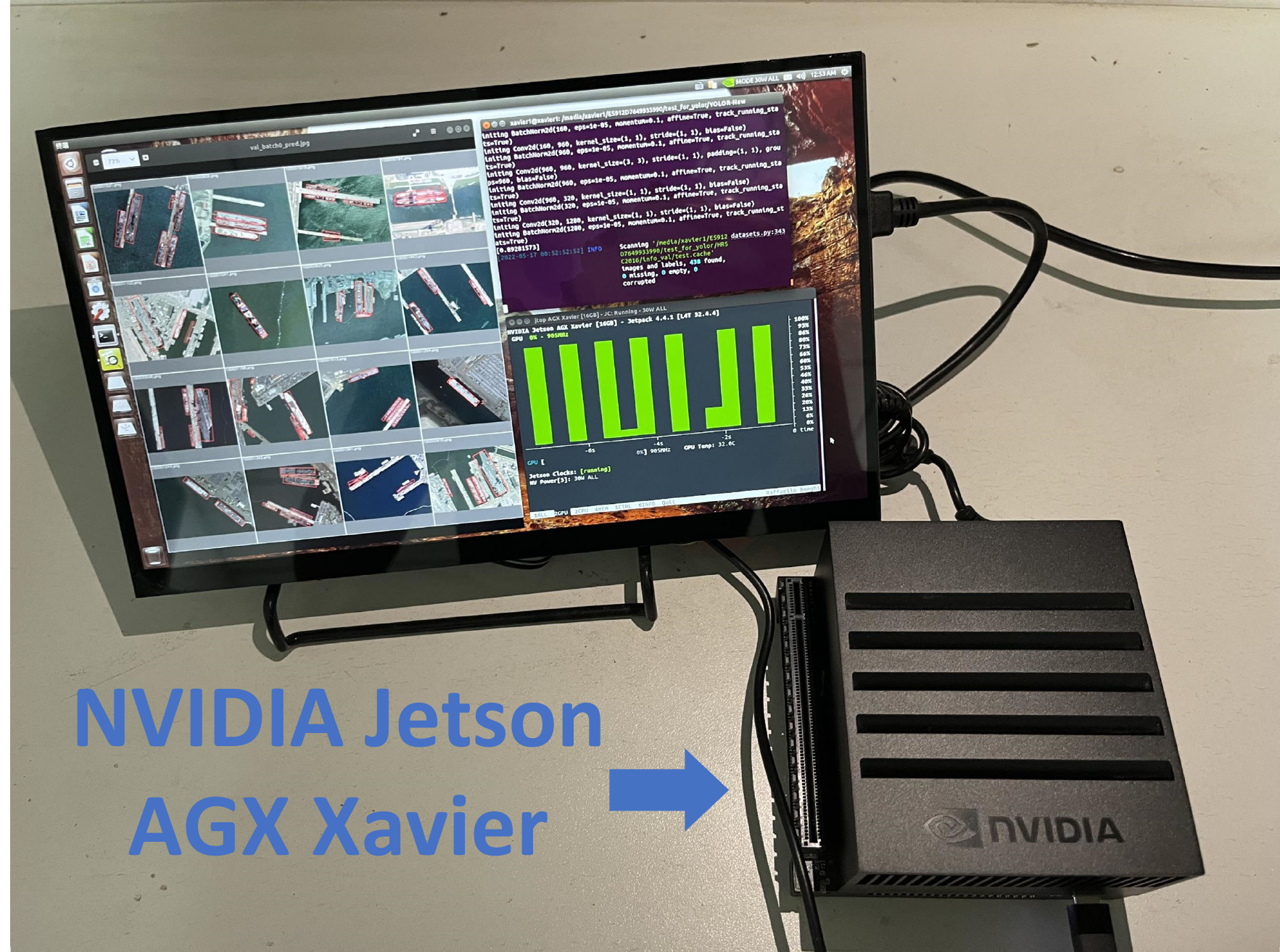}
        \caption{Visualization results of baseline+MGAR method on the DOTA dataset.}
        \label{fig6}
\end{figure}
\begin{table}[htbp]
        \centering
        \caption{Evaluation of different methods with lightweight backbone MobileNetv2 under mAP(\%)(VOC12)on the HRSC2016 dataset}
        \begin{threeparttable}
        \label{table12}
                \setlength{\tabcolsep}{0.4mm}{
                        \renewcommand\arraystretch{1.5}{


\textcolor{black}{
\begin{tabular}{l|c|c|c|c|c}
\hline\hline
Methods              & Backbone     & mAP$_{50}$     & Speed(FPS)     & FLOPs(G)       & Param.(M)      \\
\hline
baseline+CSL         & MobileNetv2 & 89.54          & 17.68          & 9.634          & 9.21           \\
baseline+DCL$^*$     & MobileNetv2 & 89.43          & 20.01          & 8.924          & 8.42           \\
baseline+MGAR$^\dag$ & MobileNetv2 & \textbf{89.67} & \textbf{21.09} & \textbf{8.915} & \textbf{8.41}  \\
\hline\hline
\end{tabular}
}
                        \begin{tablenotes}
                        \footnotesize
                        \item {\textcolor{black}{Note: The unit G is Giga, which represents $1\times10^{9}$. The unit M represents $1\times10^{6}$. Speed is the speed on NVIDIA Jetson AGX Xavier. The speed (average of 10 tests) includes the network inference speed with post-processing. The input size of image for network is 800$\times$800 pixels. $^*$ means $C_{\theta}=64$ and use Gray-Code. $^{\dag}$ means $C_{\theta}=3$. }}
                        \end{tablenotes}
                        }}
                \end{threeparttable}
\end{table}
\subsection{Lightweight Deployment}
To verify the deployment advantages of our method on lightweight devices, we
replace the backbone of the baseline with the lightweight CNN MobileNetv2 \cite {48080},
and test it on the embedded device NVIDIA Jetson AGX Xavier, as shown in Fig. \ref{fig6}.
This experiment is conducted on the HRSC2016 dataset.
The results are listed in Table \ref{table12}.
The proposed MGAR achieves the highest accuracy, and is less than CSL and DCL in terms of FLOPs and the number of parameters, with theoretically lower computational complexity, less storage consumption, and the fastest speed in the actual test.
\textcolor{black}{MGAR is nearly 3.4 FPS and 1 FPS faster than CSL and DCL respectively on the embedded device.}
Speed and accuracy results demonstrate that the proposed MGAR has excellent advantages in lightweight embedded deployment.

\section{Conclusions}
In this paper, we proposed a novel angle representation method MGAR for AOOD, which combines CAC and FAR.
The CAC eliminates the angle ambiguity introduced by previous regression-based methods, reduces the prediction layer thickness, and improves model efficiency.
The FAR refines angle prediction, which brings more accurate prediction for objects with large aspect ratios and further reduces computational consumption.
Besides, the IFL was designed to help regress angle better and more stably.
In particular, the hyperparameter introduced by MGAR is insensitive and robust to different datasets, saving extra training time.
We implemented MGAR method on the improved single-stage method YOLOv3. Extensive experiments on the HRSC2016, DOSR, UCAS-AOD, DIOR-R, and DOTA datasets demonstrated that the proposed method has excellent performance in both accuracy and speed.
We have also conducted experiments on the embedded device, and the results indicated that the proposed MGAR is very friendly for lightweight deployment and has application value.

The code is available at https://github.com/haohaolalahao.

\small
\bibliographystyle{IEEEtran}
\bibliography{MGAR.bib}

\begin{IEEEbiography}[{\includegraphics[width=1in,height=1.25in,clip,keepaspectratio]{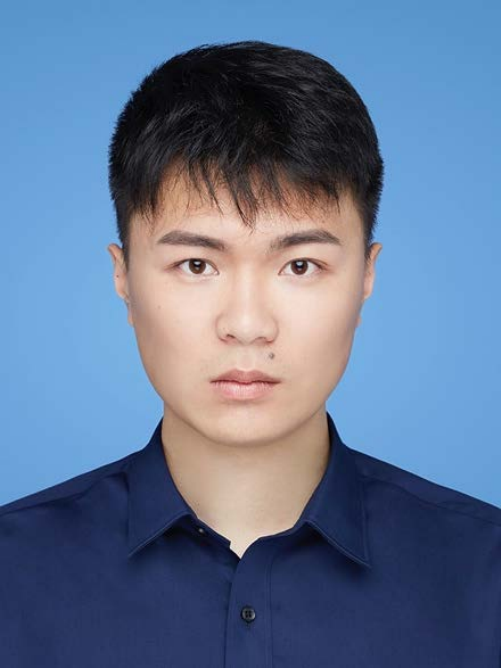}}]{Hao Wang}
	received the B.E. degree from Beijing Institute of Technology. Beijing, China, in 2020. He is currently pursuing the M.S. degree with the School of Information and Electronics Beijing Institute of Technology, Beijing, China. His research interest includes object detection and its applications
\end{IEEEbiography}

\begin{IEEEbiography}[{\includegraphics[width=1in,height=1.25in,clip,keepaspectratio]{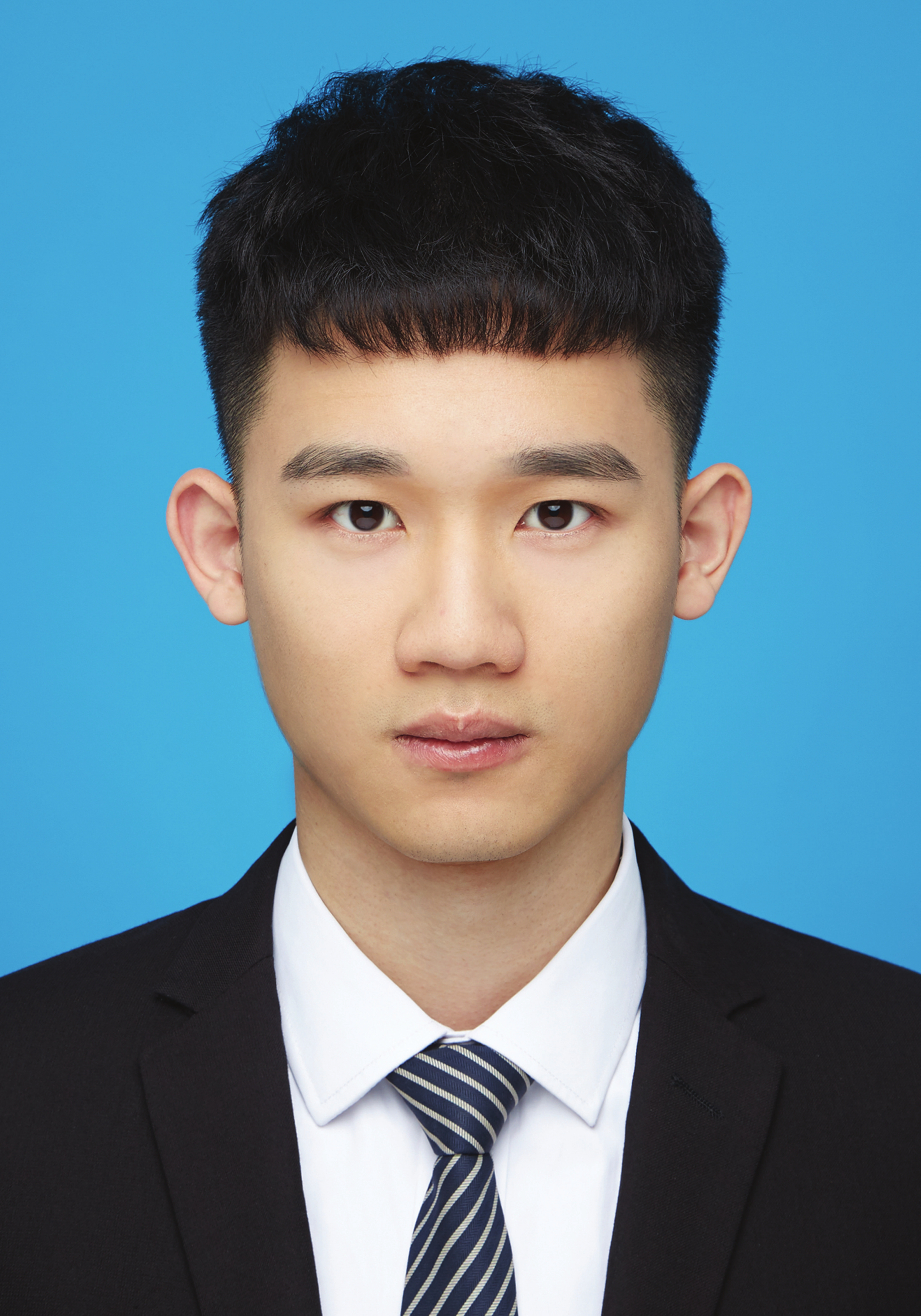}}]{Zhanchao Huang}
	received the B.E. degree and M.S. degree from Beijing University of Chemical
	Technology, Beijing, China, in 2016 and 2019, respectively. He is currently pursuing the Ph.D. degree with the School of Information and Electronics, Beijing Institute of Technology, Beijing, China. His research interests include object detection and its applications.
\end{IEEEbiography}

\begin{IEEEbiography}[{\includegraphics[width=1in,height=1.25in,clip,keepaspectratio]{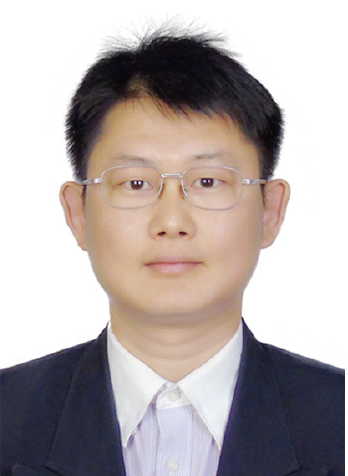}}]{Wei Li}
	(S'11, M'13, SM'16)
	received the B.E.\ degree in telecommunications engineering from Xidian University, Xi'an, China, in 2007, the M.S.\ degree in information science and technology from Sun Yat-Sen University, Guangzhou, China, in 2009, and the Ph.D.\ degree in electrical and computer engineering from Mississippi State University, Starkville, MS, USA, in 2012.
	Subsequently, he spent 1 year as a Postdoctoral Researcher at the University of California, Davis, CA, USA.

	He was a Professor with the College of Information Science
	and Technology at Beijing University of Chemical Technology from 2013 to 2019. He is currently a professor with the School of Information and Electronics, Beijing Institute of Technology. His research interests include hyperspectral image analysis, pattern recognition, and data compression.

	He is currently serving as an Associate Editor for the IEEE Signal Processing Letters and the IEEE Journal of Selected Topics in Applied Earth Observations and Remote Sensing (JSTARS). He has served as Guest Editor for special issue of Journal of Real-Time Image Processing, Remote Sensing, and IEEE JSTARS. He received the 2015 Best Reviewer award from IEEE Geoscience and Remote Sensing Society (GRSS) for his service for IEEE JSTARS and the Outstanding Paper award at IEEE International Workshop on Hyperspectral Image and Signal Processing: Evolution in Remote Sensing (Whispers), 2019.
\end{IEEEbiography}

\end{document}